\pdfoutput=1

\documentclass[11pt]{article}

\usepackage[final]{acl}

\usepackage{times}
\usepackage{latexsym}

\usepackage[T1]{fontenc}

\usepackage[utf8]{inputenc}

\usepackage{microtype}

\usepackage{inconsolata}
\usepackage{amsmath}
\usepackage{graphicx}
\usepackage{hyperref}
\usepackage{url}
\usepackage{booktabs}
\usepackage{graphicx}
\usepackage{multirow}

\usepackage{makecell}
\usepackage{caption}
\usepackage{float}
\usepackage{subcaption}
\usepackage{wrapfig}
\usepackage{color, xcolor}

\usepackage{url}
\usepackage{amssymb}

\title{MAg$\ddot{\text{I}}$C: Investigation of Large Language Model Powered \underline{\textbf{M}}ulti-\underline{\textbf{A}}gent in Co\underline{\textbf{g}}nition, Adaptab\underline{\textbf{i}}lity, Rat\underline{i}onality and \underline{\textbf{C}}ollaboration}

\author{
  Lin Xu$^{1}$\thanks{Lin Xu, \href{cathyxl2016@gmail.com}{cathyxl2016@gmail.com}. Lin Xu and Zhiyuan Hu contribute equally in this work. } \quad Zhiyuan Hu$^{1}$ \quad Daquan Zhou$^{2}$\footnotemark[2] \quad Hongyu Ren$^{3}$ \\
  \textbf{Zhen Dong}$^{4}$\thanks{Corresponding authors: \href{zhoudaquan21@gmail.com}{zhoudaquan21@gmail.com}, \href{zhendong@berkeley.edu}{zhendong@berkeley.edu}}\footnotemark[2] \quad \textbf{Kurt Keutzer}$^{4}$ \quad \textbf{See-Kiong Ng}$^{1}$ \quad \textbf{Jiashi Feng}$^{2}$ \\
  $^1$ National University of Singapore \quad
  $^2$ ByteDance \\
  $^3$ Stanford University \quad
  $^4$ UC Berkeley \\
}

\begin{document}
\maketitle
\begin{abstract}
Large Language Models (LLMs) have significantly advanced natural language processing, demonstrating exceptional reasoning, tool usage, and memory capabilities. As their applications expand into multi-agent environments, there arises a need for a comprehensive evaluation framework that captures LLMs' reasoning, planning, collaboration, and other social abilities. This work introduces a novel competition-based benchmark framework specifically designed to assess LLMs within multi-agent settings, providing quantitative metrics to evaluate their judgment, reasoning, deception, self-awareness, cooperation, coordination, and rationality.
We utilize two social deduction games alongside three game-theory scenarios to create diverse environments.
Our frame is fortified with the probabilistic graphic modeling (PGM) method, enhancing the LLMs' capabilities in navigating complex social and cognitive dimensions. We evaluate seven LLMs, quantitatively highlighting a significant capability gap of over threefold between the strongest, GPT o1, and the weakest, Llama-2-70B. It also confirms that our PGM enhancement boosts the abilities of all selected models by an average of 37\%.
Our data and code can be found here \url{https://github.com/cathyxl/MAgIC}.

\end{abstract}

\begin{figure}[!th]
    \centering
    \includegraphics[width=0.9\linewidth]{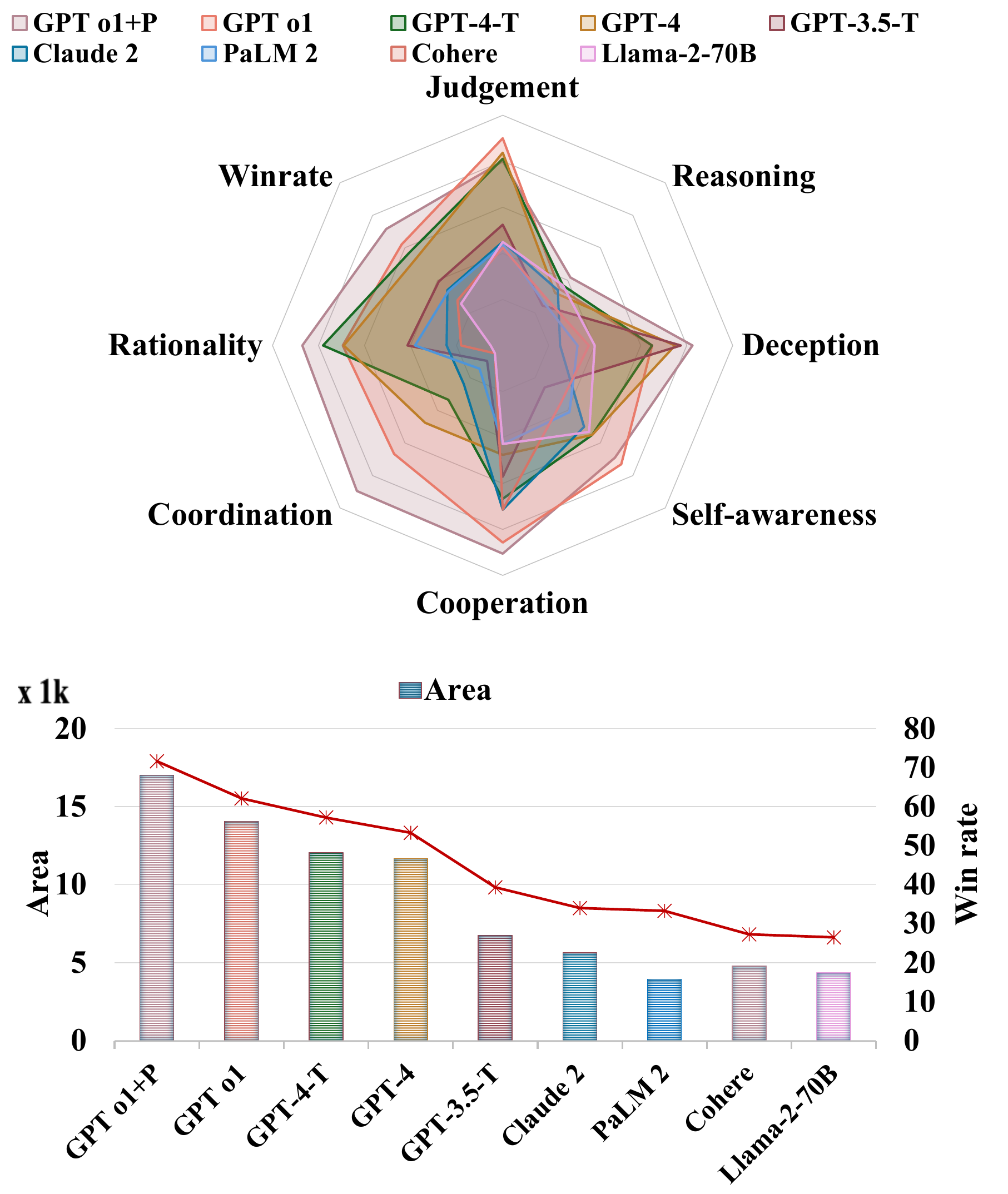}
    \vspace{-3mm}
    \caption{The radar chart depicts LLMs' performance on 7 metrics, with ``-T'' for ``-turbo'' and ``+P'' for ``+PGM''. The bar chart displays the polygons' areas, and the red line indicates average game-winning rates. Larger areas correlate with higher winning rates, validating the effectiveness of the proposed metrics for assessing LLMs' capabilities. For more information, refer to Sec. \ref{sec:experiments}.} 
    \vspace{-10pt}
    \label{fig:tease}
\end{figure}

\section{Introduction}

\label{sec:intro}
Large language models (LLMs), particularly ChatGPT and GPT-4 \cite{openai2023gpt4}, have showcased impressive understanding and generation capabilities. Beyond these fundamental abilities, LLMs also demonstrate promising capabilities in anthropic areas such as reasoning \cite{wei2022chain}, planning \cite{hao2023reasoning}, tool usage \cite{schick2023toolformer}, and memorization \cite{shinn2023reflexion}.
There is an increasing interest in investigating LLMs' behaviors as agents in single- or multiple-agent systems. Noteworthy examples include Generative Agents \cite{park2023generative}, Camel \cite{li2023camel}, Auto-GPT \cite{richards2023auto}, and Voyager \cite{wang2023voyager}.

Meanwhile, quantitative assessment of LLMs as agents is crucial for their advancement.  Recent benchmarks, such as \citet{liu2023agentbench}, evaluate LLM-as-Agent in multi-turn contexts, while concurrent work by \citet{wu2023smartplay} tests them in games requiring reasoning and planning. However, these studies focus on understanding and reasoning in environments, overlooking true interaction capabilities in multi-agent systems. Other research, including \citet{agashe2023evaluating} social aspects on coordination, and \citet{fu2023improving} on bargaining, explores specific skills in multi-agent scenarios. While these studies provide useful insights into LLMs' certain capabilities, their scope is limited and lacks quantitative metrics.

We have observed three key characteristics in interactive multi-agent systems. 
(1) Agents in these systems often operate within the confines of their local perspectives. However, making wise decisions typically necessitates a good understanding of global information. To overcome this limitation, agents must adeptly discern contexts and reason about the roles or plans of other agents. 
(2) Contexts are inherently dynamic in multi-agent systems due to the dependent nature of agents' decisions. Success hinges on the ability to swiftly adapt strategies in response to evolving contexts. 
(3) Collaboration and competition are inevitable when multiple agents try to solve tasks together. The ability to promote cooperation while preserving self-interest is often the ultimate goal of multi-agent systems.

Inspired by the above characteristics, we first propose a competition-based benchmark to evaluate the abilities of LLMs as agents by competing with a fixed type of LLM. Besides, seven quantitative metrics from the competitions are proposed to measure the essential capabilities of LLMs~\cite{wooldridge2009introduction, minsky1988society}. We define these capabilities from four aspects: cognition, adaptability, rationality, and collaboration:
(1) \textbf{Judgment} and \textbf{reasoning} form the core \underline{\textit{\textbf{cognition}}} of agents, crucial for \textit{accurate information estimation} in uncertain scenarios. Judgment evaluates the ratio of the final correct decisions. Reasoning measures the ability to logically analyze other agents' roles and strategy formulation, thus guiding agents to make correct decisions in uncertainty. 
(2) \textbf{Self-awareness} and \textbf{deception} are key to \textit{enhanced \underline{\textbf{adaptability}}} in agents, vital for multi-agent system. Self-awareness is an assessment of agents' understanding of their capabilities and roles, ensuring the consistency of behaviors towards the target. Deception enables agents to subtly manipulate information in competitive settings, influencing other agents' decisions and gaining advantages in social interactions.
(3) \underline{\textit{\textbf{Rationality}}} serves as a metric to gauge the \textit{efficiency} of an agent's behavior. It directs agents toward making decisions with the aim of optimizing their benefits by considering the potential actions of other agents rather than resorting to impulsive or uninformed actions.
(4) \textbf{Cooperation} and \textbf{coordination} are two facets of \underline{\textit{\textbf{collaboration}}}, essential for effective teamwork in multi-agent systems. Cooperation measures communication and agreeability. Coordination indicates collaboration facilitation.

In light of the essential abilities, we further propose a method to enhance LLMs as agents by integrating Bayesian statistical foundations. This novel approach intertwines the Probabilistic Graphical Model(PGM)~\cite{koller2009probabilistic} with LLMs, thereby amplifying their capacity to comprehend intricate scenarios and enabling more informed and strategic decision-making in multi-agent environments.

In summary, our contributions are as follows:

\begin{itemize}
    \item We first propose a competition-based benchmark environment for LLM-powered multi-agent systems by collecting over 100 cases in 5 scenarios and designing 7 metrics to evaluate the critical abilities in multi-agent systems.

    \item We measure 7 LLMs with our benchmark. The results indicate that GPT o1, GPT-4, and GPT-3.5 remain the superior performers, followed by other commercial LLMs - PaLM 2, Claude 2, and Cohere. Different large language models (LLMs) exhibit varying performance levels across different evaluation dimensions and possess distinct characteristics. For instance, GPT-o1 is more discernible with a good judgment score, GPT-4 tends to be more rational, whereas GPT-3.5 is generally more cooperative, as shown in~\autoref{fig:tease}.
    
    \item We design a PGM-aware agent that integrates LLMs and symbolic reasoning to fortify itself in multi-agent systems. PGM-aware agents outperform their vanilla versions by 37\% on average over these abilities. As shown in~\autoref{fig:tease}, GPT o1+PGM has achieved impressive improvement over the original GPT o1.
\end{itemize}

\section{Related Work}
\textbf{Emergent Capabilities of LLMs.}
Beyond their core functions, LLMs have shown diverse emergent abilities like reasoning, planning, memory and so on. Recent works like Chain of Thought \cite{wei2022chain}, Tree of Thought \cite{yao2023tree}, Graph of Thought \cite{yao2023beyond, besta2023graph}, and ReAct \cite{yao2210react} improve LLM reasoning. API-bank \cite{li2023api} benchmarks tool-augmented LLMs, with ToolLLM \cite{qin2023toolllm} providing a framework using tools. Reflexion \cite{shinn2023reflexion} enhances LLM decision-making, while Phelps investigates economic goal-like behavior \cite{phelps2023investigating}. 

\noindent\textbf{LLMs-Powered Agents.}
Advancements in LLMs ignite agents dealing with intricate tasks~\cite{richards2023auto, li2023camel,wang2023voyager} and more complex scenarios involving multiple agents~\cite{park2023generative}. Auto-GPT \cite{richards2023auto} demonstrates GPT-4's capabilities in achieving goals through chained thoughts. Generative Agents \cite{park2023generative} describes a sandbox with 25 AI agents simulating human actions, recording experiences for deeper self-awareness. 
Meanwhile, recent and concurrent studies conduct benchmarking for LLM-powered agents. Some studies~\cite{liu2023agentbench,wu2023smartplay,gioacchini2024agentquest} evaluate the capabilities of LLM-powered single agents within games or real-life environments. Others study LLMs' social abilities in multi-agent systems. \citet{agashe2023evaluating} explores coordination ability. \citet{abdelnabi2023llm} assess the deliberation ability of LLMs in negotiation games.
A concurrent work~\cite{huang2024far} tests LLMs in 8 game theory scenarios, analyzing the abilities of one LLM in multi-agent playing. However, these works still lack quantifiable measurements for social abilities. 
%

\section{Benchmark}
\begin{figure}[!th]
    \centering
    \includegraphics[width=0.98\linewidth]{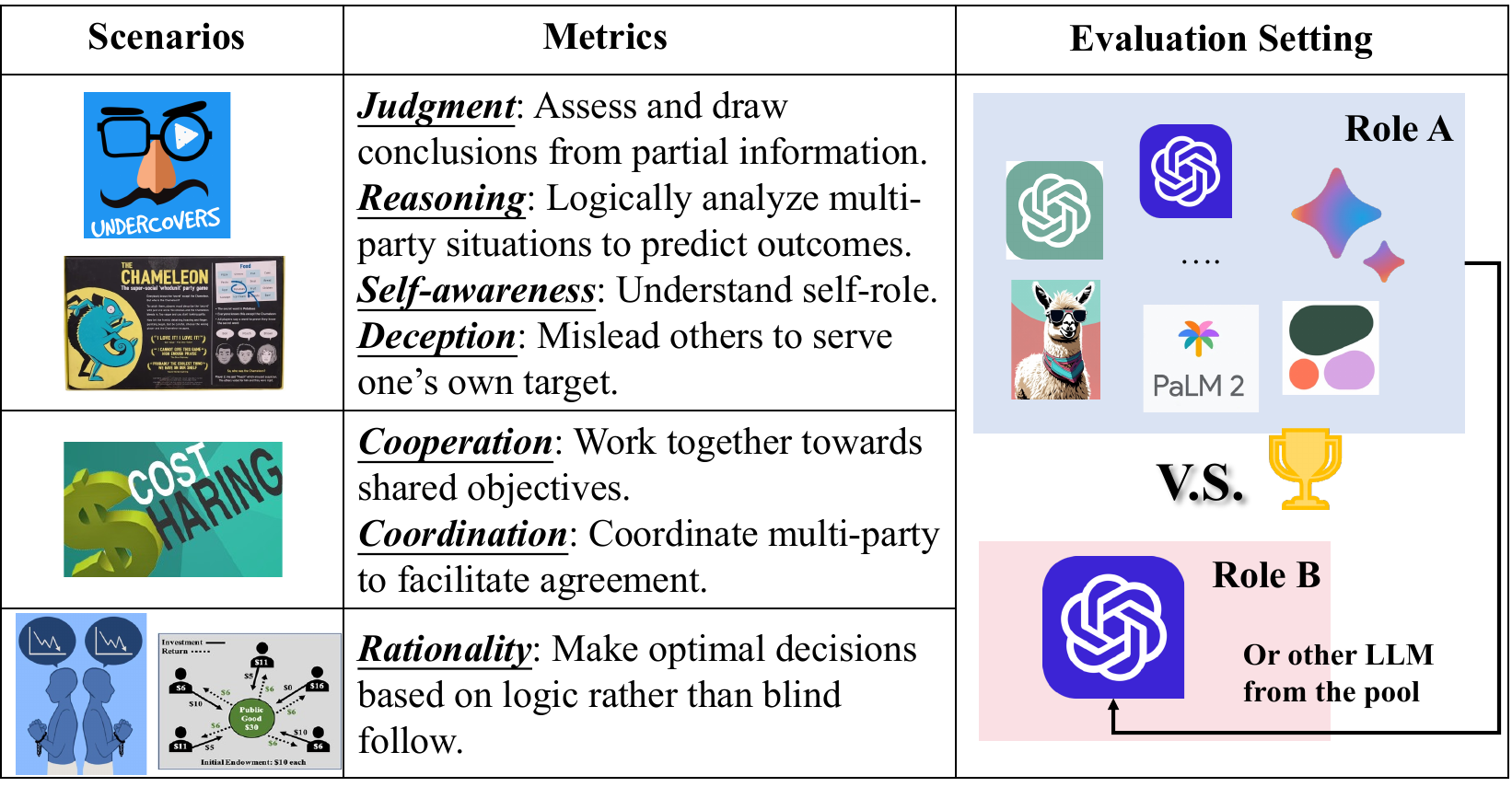}
    \vspace{-10pt}
    \caption{Overview of evaluation setting, scenarios, and proposed metrics.}
    \vspace{-7pt}
    \label{fig:benchmark}
\end{figure}
\noindent We propose to measure the abilities of various LLMs by putting them into competitions of multi-agent scenarios. In this way, we measure the genuine capabilities of LLMs when interacting with multiple other agents. To achieve this, we have constructed a comprehensive benchmark that incorporates various competition settings and meticulously designed metrics for each scenario. The detailed constitutes are illustrated in ~\autoref{fig:benchmark}.

\subsection{Scenarios} 

As mentioned in Sec.~\ref{sec:intro}, the evaluation of agents in multi-agent systems revolves around crucial attributes such as cognition, adaptability, rationality, and collaboration. We select scenarios according to two scenarios: 1) complex enough, requiring agents to exhibit good global comprehension and information manipulation ability; 2) emphasizing both collaboration and rationality to balance both global and self-interests.
In the game of the social deduction games Chameleon and Undercover, quickly grasping global information and taking clever actions are the keys to winning the game. Thus, we mainly measure the cognition and adaptability in these two scenarios. Moving to game theory scenarios, which require the agent to make optimal decisions based on the given premise~\cite{myerson1991game}, they are more apt for reflecting rationality and collaboration. As such, we center our evaluation on these latter two attributes in the context of three game theory scenarios. 

\noindent\textbf{Chameleon} is a social deduction game where players are either a chameleon or a non-chameleon. Non-chameleon players give clues about a secret word. The chameleon player tries to blend in without knowing the word. Non-chameleons aim to expose the chameleon without revealing the word. 

\noindent\textbf{Undercover}, as a similar game, divides players into civilians and undercovers. The word for undercovers is different from civilians. Players explore their roles by telling from their own and other's clues. Civilians need to find undercover, while undercovers should hide themselves. 

\noindent\textbf{Cost Sharing} involves multiple parties dividing costs based on their usage of a shared resource. These parties need to propose and negotiate cost allocation solutions. Each party are expected to ensure fairness to achieve unanimous agreement and meanwhile reduce their own cost to realize largest interest.

\noindent\textbf{Multi-turn Prisoner's Dilemma} extends the classic Prisoner's Dilemma to a multi-round three-player version. Each participant decides to cooperate or defect in every round, and the scores are determined by collective choices. For example, if only one player defect while others cooperate, the betraying player will get highest score. The game tests players' ability to strategize, foster trust, and navigate group decision-making. The player with the highest total score at the end of the game is declared the winner. 

\noindent\textbf{Public Good} explores similar strategies in Prisoner's Dilemma. Players are given fixed initial resources. They can decide how much to invest to a common pool at each round. The total investment from all the players is then multiplied and distributed to each player evenly. The winner is the player possessing the most resources at the end.

\subsection{Competition Settings}
We propose a competition-based evaluation to ensure genuine multi-agent interactions and comparability among different LLMs. In this setting, different LLMs(referred to as challenger LLMs) challenge the same defender agents (powered by a fixed LLM), in the same game settings. Then their capabilities are evaluated based on the meaningful intermediate game results, and the winning rates over defender agents. LLM with higher wining rates are more capable, based on which we can rank the ability of different LLMs. As shown by Evaluation Setting in ~\autoref{fig:benchmark}, GPT-4 is used as the defender LLM, and other LLMs challenge to be the champion. We've gathered a collection of cases for each scenario mentioned above. The detailed competition procedures and collection process are in~\ref {append:competition_setting}. 

\subsection{Evaluation Metrics}
In assessing the seven capabilities within a multi-agent system, we define the following metrics. First, we average the win rates of all the roles the challenger LLM plays in all scenarios as an overall score, Win Rate. 

\noindent \textbf{Win Rate} is a straightforward indicator of the success of an LLM in all proposed scenarios.
    \begin{equation}
        w_r=\frac{1}{|\mathcal{S}|}\sum_{s_i\in\mathcal{S}}w_{s_i}
    \end{equation}
Here $\mathcal{S}$ is the set of roles the challenger LLM plays in all the scenarios. In Chameleon and Undercover, the challenger LLM play the Chameleon, Non-Chameleons, Undercovers, and Civilians respectively. As for game theory scenarios, the challenger LLM plays one of the players. Thus, the length of $\mathcal{S}$ is 7 in our benchmark. For each role, we have defined the criteria for winning and denote the win rate as $w_{s_i}, s_i \in \mathcal{S}$. The detailed definitions for winning rates of all roles can be found in~\ref{append:win_rate_define}.
    
\begin{figure*}[h]
    \centering
    \includegraphics[width=0.9\linewidth]{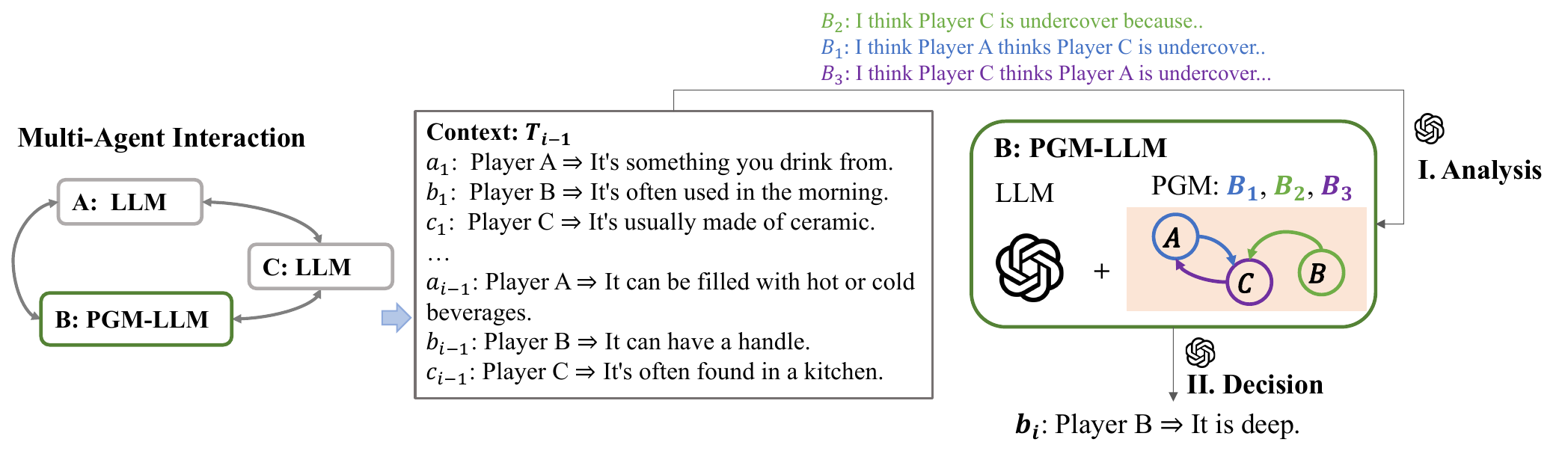}
    \vspace{-10pt}
    \caption{A Decision process of the PGM-aware agent. This example involves an undercover game where the PGM-Aware agent B believes that agent C is the undercover. Consequently, B decides to respond with "It is deep," which better describes the features of the word "cup" rather than the undercover word "mug". }
    \label{fig:pgm_agent}
    \vspace{-10pt}
\end{figure*}

\noindent \textbf{Judgement} measures the final understanding of the global information, essential for assessing LLM's ability to distinguish other players' identities based on partial information. In our benchmark, we use the correct vote ratio in Chameleon and Undercover to indicate the ability, formulated as:
\begin{equation}
S_{J} = n_{cv} / n_v
\end{equation}
, where $n_{cv}$ and $n_v$ are the number of correct votes and total votes when the challenger LLM are playing civilians and non-chameleons.
    
\noindent \textbf{Reasoning} evaluates the correctness of agents' analysis about multiple parties, which often requires multi-hop logical reasoning based on the global settings and partial information from other players. 
We ask each player deduces other players' roles and also predict a step further about other players' deductions. By comparing these deductions with the gold situations and other players' true subjective deductions, we can decide whether their rightfulness. We denote number of these two types of deductions as $n_{\text{gold}}$ and $n_{\text{inter}}$. The number of correct deductions as $n_{\text{c\_gold}}$ and $n_{\text{c\_inter}}$. The Reasoning is defined as:
    \begin{equation}
    S_{R} = (n_{\text{c\_gold}} + n_{\text{c\_inter}})/(n_{\text{gold}} + n_{\text{inter}})
    \end{equation}    
\noindent \textbf{Deception} presents an agent's capability to deceive others to serve their goal. We measure this by the ratio of successful deceptions. In detail, our benchmark calculates the ability as the ratio for chameleon/undercover's successful blending or causing incorrect secret word guesses, denoted as:
    \begin{equation}
    S_{D} = n_{\text{wuc}}/n_{\text{uc}} + \lambda (n_{\text{wcg}}/n_{\text{cg}})
    \end{equation}
    where $n_{\text{wuc}}$ and $n_{\text{uc}}$ are the win count and total count of games when the LLM plays chameleon and undercover, $n_{\text{wcg}}$ is the number of incorrect code guesses, and $n_{\text{cg}}$ is the total number of code guesses. Here, we assign a weight $\lambda=0.25$ due to not all the games trigger code guesses. 

\noindent \textbf{Self-Awareness} measures correct role identification, ensuring correct and consistent behavior following their own roles. 
    \begin{equation}
    S_{\text{self}} = \mu (n_{\text{{crc}}} / n_{\text{rc}})+ n_{\text{{cru}}} / n_{\text{ru}}
    \end{equation}
    where $n_{\text{crc}}$, $n_{\text{rc}}$ is the number of correct and the total number of role identifications in chameleon, and $n_{\text{cru}}$, $n_{\text{ru}}$. $\mu=0.6$ is used because it is much easier to identify roles in a chameleon game. 
    
\noindent \textbf{Cooperation}. The ability to cooperate with other players and achieve a common goal. Our benchmark measures it in cost-sharing games, showcasing an agent's effect on the collective efficacy of the system.
    \begin{equation}
    S_{\text{collab}} = n_{\text{wcs}}/n_{\text{cs}} 
    \end{equation}
    where $n_{\text{wcs}}$ and $n_{\text{cs}}$ are the number of successful and the total number of cost-sharing games.
    
\noindent \textbf{Coordination} measures how the LLM contributes to successful collaboration by providing constructive proposals. We formulate the metric in our benchmark as follows: 
    \begin{equation}
    S_{\text{coord}} = n_{\text{\text{pcs}}}/n_{\text{wcs}}
    \end{equation}
    where $n_{\text{pcs}}$ is the number of successful collaborations proposed by the challenger LLM in the cost-sharing games.
    
\noindent \textbf{Rationality} captures the agents' ability to act rationally to optimize their own interests according to the rules of the game theory scenarios~\cite{myerson1991game}. Suppose there are $\mathcal{T}_{\text{pd}}$,  $\mathcal{T}_{\text{pg}}$ rounds in each competition for Prisoner's Dilemma and Public Good. The Rationality is defined as:
    \begin{equation}
    S_{R} = \frac{n_b}{n_{\text{pd}}*\mathcal{T}_{\text{pd}}} + \frac{n_{\text{li}}}{n_{\text{pg}}*\mathcal{T}_{\text{pg}}}
    \end{equation}
    where $n_b$ is the round of betray decisions, $n_{\text{li}}$ is the round of decisions where the challenger LLM invests the least in the common pool, $n_{\text{pd}}$ and $n_{\text{pg}}$ are the number of prisoner's dilemma competitions, and the number of public good competitions, respectively.

\section{PGM-Aware Agent}

In AI, Bayesian methods embody symbolism, while large language models (LLMs) exemplify connectionism. Despite their individual strengths, effectively combining these approaches remains a challenge. LLMs are proficient in complex language tasks but still struggle with ambiguous relationships and causal reasoning. This shortcoming is especially evident in multi-agent scenarios requiring complex inferential analysis. To address this, we propose integrating Probabilistic Graphical Models (PGMs), classic Bayesian tools adept at depicting dependencies between random variables, to enhance LLMs' analytical and inferential capabilities.

\subsection{PGM Structure}
We leverage PGM to depict intricate dependency relationships among all agents, thereby augmenting the LLMs' comprehension of the global information. This heightened understanding can subsequently facilitate informed actions/decisions. The PGM should be comprehensive and thorough to ensure wise decision-making for an agent. For instance, considering the prisoner's dilemma scenario, before deciding to defect or cooperate, it is crucial to anticipate whether others might defect or cooperate and, from others' perspectives, how you will decide. If you anticipate that other players cooperate and they expect the same from you, but you choose to defect, it can lead to a significant advantage for you. As a result, We design the PGM structure in a two-hop understanding mechanism in which the agent analyzes from its own perspective and perspective when it stands in other agents' shoes. This is highly relevant to the psychological concept of Theory of Mind(ToM)~\cite{baker2011bayesian, oguntola2023theory}, which is the capacity to comprehend human actions by predicting their unknown beliefs and desires. We employ the Probabilistic Graphical Model (PGM) to provide a general formalization of this concept. PGM uses graphs to illustrate the conditional dependencies between random variables~\cite{koller2009probabilistic}, making it particularly suitable for understanding interactions among multiple players.

Formally, as shown in ~\autoref{fig:pgm_agent}, suppose there are three players A, B, and C, in one game and they've played the game for $i-1$ turns and formed the context $T_{i-1}=\{a_1, b_1,..., a_{i-1},b_{i-1},c_{i-1},\}$. Here $a_*$, $b_*$, and $c_*$ are the decisions from Players A, B, and C, respectively. As a PGM-aware player, B manages three distinct random variables, denoted as $B_1$, $B_2$, and $B_3$, representing B's interpretations of the global status from A, B, and C's perspective. We obtain the estimation for these random variables by prompting LLMs through different prompts as listed in ~\ref{apped:game_rule_and_prompt}, $\mathcal{P}^{\text{pgm}}_j, j\in [1,2,3]$:
\begin{equation}
    P(B_j) = \text{LLM}(B_j|\mathcal{P}^{\text{pgm}}_j, T_{i-1})
\end{equation}
In designing the Probabilistic Graphical Model (PGM), we have opted not to limit its representation solely to numerical probabilities. Instead, we also incorporate text-represented probabilities, acknowledging the text-based input and output nature of large language models (LLMs). The primary purpose of the PGM is to structure the multi-party, multi-hop understanding mechanisms within multi-agent systems.

\begin{table*}[t]
\centering \small
\footnotesize
\resizebox{0.8\textwidth}{!}{
\begin{tabular}{l|c|ccccccccc}
    \cmidrule[\heavyrulewidth]{1-9}
    & \textbf{Win Rate}& \textbf{Judge.} &  \textbf{Reason.}  &  \textbf{Decept.}  &   \textbf{Self-aware.} & \textbf{Cooper.}  & \textbf{Coord.}&  \textbf{Rational.} \\
    \cmidrule(lr){1-9}

    GPT o1 & \textbf{62.1} & \textbf{90.0} & 34.8  & 65.0  & \textbf{73.0} & \textbf{85.7} &\textbf{66.7} & 69.5 \\
    GPT-4-turbo      &57.2 &81.2 &\textbf{37.0} &65.0 &55.0 &66.7 &33.4 &\textbf{78.1} \\
    GPT-4            & 53.3 & 83.8 & 32.3 & 75.0 & 55.0 & 47.6 & 47.6  & 69.0 & \\
    GPT-3.5-turbo       & 39.3 & 52.5 & 24.5 & \textbf{77.5} & 25.9 & 57.1 & 9.50 & 41.4 \\

    Claude 2       &34.0 & 45.0 & 34.0 & 25.0 &50.0 & 71.4 &23.8 & 24.3\\

    PaLM 2       &33.3 &43.8 &25.8 & 32.5 & 41.1 & 42.9 &14.3 &38.1  \\

    Cohere &27.3 &42.5 &27.8 &37.5 &35.6 &71.4 &4.80 &18.1 \\
    
    Llama-2-70B      & 26.5 & 45.0 & 37.0 & 40.0 & 53.2 & 42.9 & 4.80 & 5.20 \\

    \cmidrule[\heavyrulewidth]{1-9}

\end{tabular}
}
\caption{Ability Measurements of Different LLMs.}
\label{table:bechmarking}
\end{table*}

\begin{figure*}[t]
    \centering
    \includegraphics[width=0.8\linewidth]{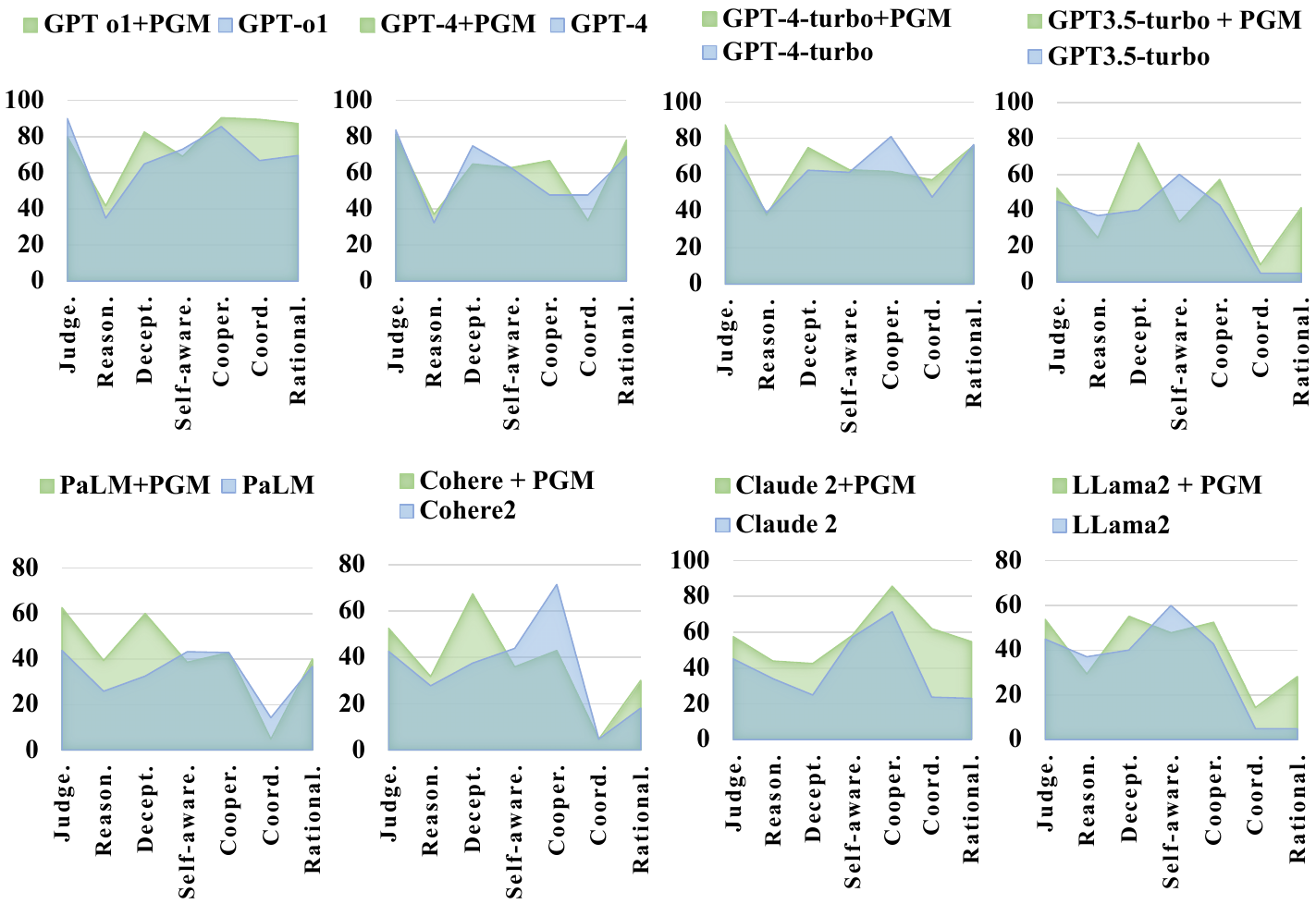}
    \caption{The comparison between PGM-aware and vanilla agents involves seven metrics. Most PGM-aware agents significantly outperform the vanilla ones in 3-4 out of the 7 abilities, with p-values lower than 0.05(t-test).}
    \label{fig:pgm_comparison}
\end{figure*}

\subsection{LLM Decision with PGM}
For the LLM agent in multi-agent, the inference process is formulated as:

\begin{equation}
\label{eq:pgm}
    P(b_i)= \text{LLM}(b_i|\mathcal{P}, T_{i-1})
\end{equation}
where $\mathcal{P}$ is the prompt to let the LLM go to the next step.
Our PGM-Aware Agent makes decisions conditioned both on the PGM and game contexts, which can be formulated as:
\begin{equation}
    P(b_i)=\text{LLM}(b_i|\mathcal{P}^{\text{decision}}, B_1,B_2,B_3,T_{i-1})
\end{equation}
where $\mathcal{P}^{\text{decision}}$ is the prompt to guide the LLM to make a decision given both PGM and context in the next step. $B_1,B_2,B_3$ are the PGM acquired in ~\autoref{eq:pgm}. We have listed the prompts used in basic LLMs and the PGM-Aware Agent in ~\ref{apped:game_rule_and_prompt}.

\section{Experiments}
\label{sec:experiments}
In experiments, we make each challenger LLM play with the same defender LLMs(GPT-4 as we used), and rank them by the wining rate. To reduce the randomness during game, we set the temperature of all participating LLMs as 0. All the code and data will be publicly released upon acceptance.
\subsection{LLM Leaderboard}
We evaluate GPT-3.5-turbo \cite{openai2023gpt3.5}, GPT-4 \cite{openai2023gpt4}, Llama-2-70B \cite{touvron2023llama}, PaLM 2 \cite{anil2023palm}, Cohere \cite{cohere} and Claude 2 \cite{Claude-2} with our benchmark. In~\autoref{fig:tease}, we clearly compare the capabilities of different LLMs. The most prominent performer is the GPT-4-turbo method, showcasing outstanding overall performance with a remarkable win rate of 57.2\%. This significantly higher win rate underscores its competitive advantage. Following closely is GPT-4, which achieves a win rate of 53.3\%, demonstrating its competitiveness. 

Furthermore, as illustrated by the radar chart in~\autoref{fig:tease} and the corresponding area calculations in the lower bar chart, GPT-4-turbo surpasses Llama-2-70B by more than threefold in overall multi-agent capabilities. Additionally, GPT-3.5-turbo also demonstrates superior performance compared to Llama-2-70B. Our evaluation of other popular commercial LLMs, including PaLM 2, Claude 2, and Cohere, shows that their multi-agent abilities fall between those of GPT-3.5-turbo and Llama-2-70B. Notably,~\autoref{fig:tease} indicates that the area sizes derived from the proposed abilities' values are directly proportional to the winning rates. This correlation validates our benchmark as an effective tool for assessing the capabilities of different LLMs.

As demonstrated in~\autoref{table:bechmarking}, we conducted a detailed comparison by evaluating metrics such as Judgment, Deception, Reasoning, and Self-Awareness within the Chameleon and Undercover scenarios. In these contexts, GPT-o1 present impressive scores in Judgment, Self-awareness, Cooperation and Coordination.  GPT-4 excelled with scores of 90\% in Judgment and 75.0\% in Deception, solidifying their superiority in these scenarios. The performance gap in reasoning abilities among the models was narrow, while deception capabilities showed significant disparities.
On the other hand, when assessing metrics related to collaboration, coordination, and rationality in game theory scenarios like Cost Sharing, Prisoner's Dilemma, and Public Good, GPT-4 and GPT-4 Turbo continued to shine. GPT-4 achieved 66.7\% in Coordination and an optimal performance of 78.1\% in Rationality. In contrast, Llama-2-70B, while lagging in overall performance with a win rate of 26.5\%, exhibited strengths in specific metrics, such as a relatively high self-awareness score of 53.2\%, surpassing GPT-3.5 Turbo's 25.9\%.

\subsection{PGM Enhancement Performance}
\label{subsec:pgm_enhance}

As shown in \autoref{fig:pgm_comparison}, the green section highlights the effectiveness of the PGM-aware approach. This enhancement is particularly pronounced in the PaLM, Claude 2, and Llama2 models, as detailed in \ref{append:pgm}. Overall, the PGM-aware method has achieved average improvements across all capabilities by a margin of 37\%, calculated by comparing the radar areas achieved by vanilla in~\autoref{fig:tease} and PGM-aware methods in~\autoref{fig:radar_pgm}. PGM-aware methods have increased the win rate in all scenarios by 6.57\%.

For each capability, as illustrated in \ref{append:pgm}, PGM-aware methods have achieved an 8.72\% increase in Judgement, confirming the method's ability to enhance analysis in LLMs. Reasoning and Deception abilities have seen improvements of approximately 5\% and 6\%, respectively. Notably, the most significant enhancements are observed in Coordination and Rationality, with improvements of 12.2\% and 13\%. We've also done significance tests(t-test) for each pair of the vanilla LLM and its PGM-aware version. Most PGM-aware agents significantly outperform their vanilla counterparts in 3 or 4 out of the 7 abilities, as shown in~\autoref{fig:pgm_comparison}, with p-values lower than 0.05. GPT-3.5-turbo is also significantly improved in rationality and deception, and GPT-4 is significantly improved in Cooperation.


\begin{figure*}[!th]
    \centering
    \includegraphics[width=0.9\linewidth]{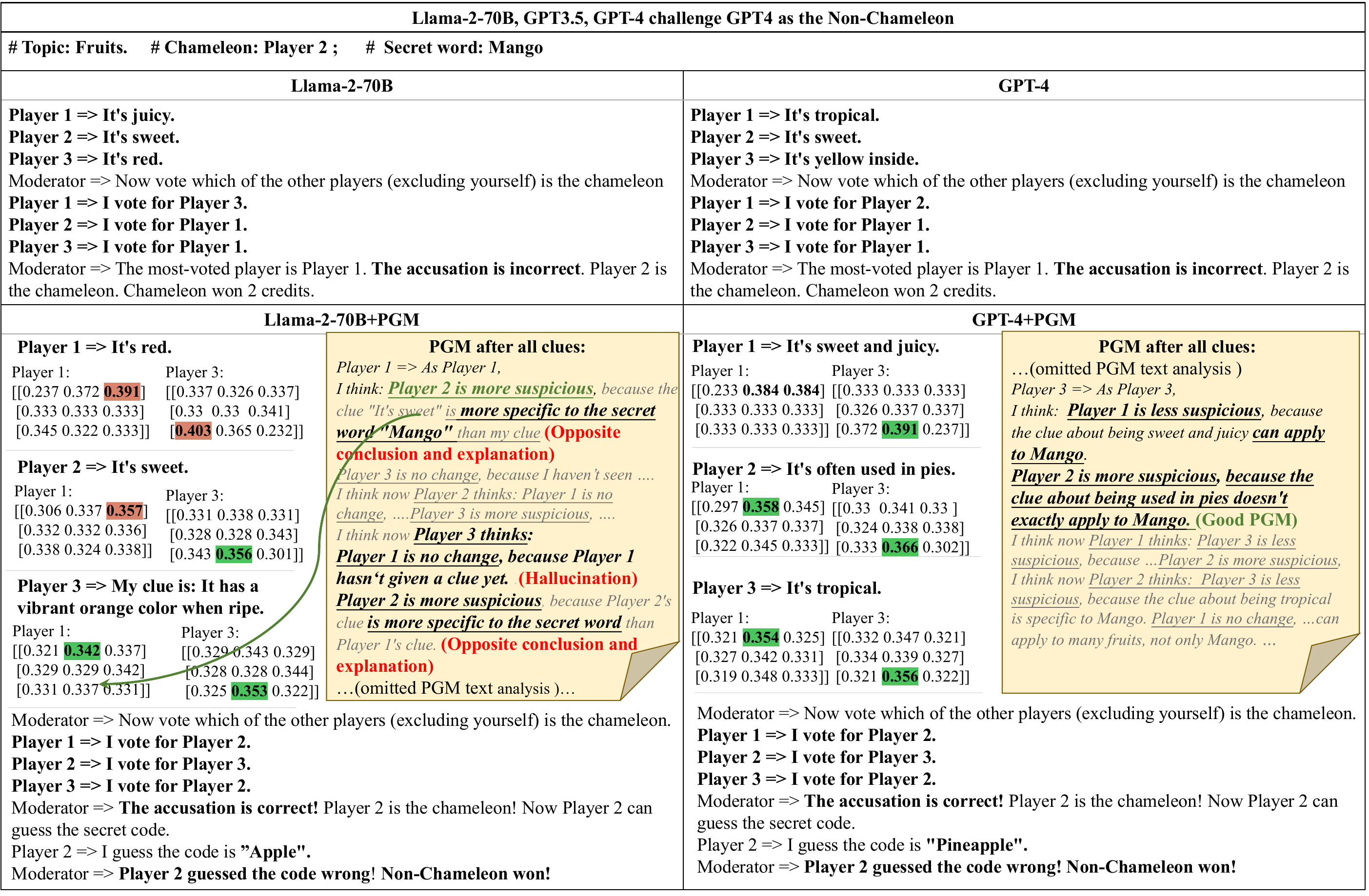}
    \caption{A case study on Chameleon, Llama-2-70B, GPT-4, and their PGM-enhanced versions. The numerical probabilities are calculated by extracting judgments in the text-based PGM and normalized into scale of 0 to 1. }
    \vspace{-10pt}
    \label{fig:chameleon_cases}
\end{figure*}
\subsection{Analysis}
\label{sec:discussion}
The above experimental results prove that the PGM-Aware agent can improve the performance of each metric to varying degrees. The discussion part explores the influence of PGMs from other aspects in each scenario.

\noindent\textbf{RQ1. How PGM of different LLMs help Judgement and Reasoning? }
In~\autoref{fig:chameleon_cases}, we provide a case of Llama-2-70B, GPT-4, and their PGM-Aware versions playing as non-chameleons versus GPT-4 as the chameleon. For Llama-2-70B and GPT-4, we can find both LLMs failed to win the game because they voted for the wrong chameleon ``Player 1''. After being equipped with PGM, both models change the game results to Non-chameleons won. If we look into the details of the game process, in the first two clues, the PGMs given by LLama-2-70B all indicate wrong chameleons, as highlighted in red in~\autoref{fig:chameleon_cases}, since ``Player 2'' is the true chameleon. In the third round, ``Player 1'' gave another right analysis which successfully changed the PGM to the right indication of the chameleon. However, if we look at the content of the analysis, we find the analysis gave opposite conclusions and explanations. For example, ``more suspicious'' is ``more specific to the secret word''. Besides, LLama-2-70B also presents some hallucinations in the game, for example, after Player 1 already gave the clue, the analysis still states ``Player 1 hasn't given a clue yet''. While GPT-4+PGM's analysis aligns the conclusions and explanations well and has no hallucinations. According to the example, we can find PGM could be helpful for models to make better judgments through clear analysis and PGM is affected by the ability of LLMs. The more powerful the model, the more accurate its judgment and reasoning. 
 
\begin{table}[!th]
\centering \small
\resizebox{\linewidth}{!}{
\begin{tabular}{lcccccc}
    \cmidrule[\heavyrulewidth]{1-7}
    \textbf{LLM} & \multicolumn{2}{c}{\textbf{Cost-Sharing}} & \multicolumn{2}{c}{\textbf{Prisoner}} & \multicolumn{2}{c}{\textbf{Public Good}} \\
    &  \textbf{WR}$\uparrow$ & \textbf{Cost}$\downarrow$ & \textbf{WR}$\uparrow$ & \textbf{Score}$\uparrow$ & \textbf{WR}$\uparrow$ &\textbf{Payback}$\uparrow$ \\
    \cmidrule(lr){1-7}
    Llama-2      & 42.8 & 37.1  & 0.0  & 6.05  & 0.0  & 139.1  \\
    Llama-2+P  & 52.4 & 37.6  & 38.5 & 9.86  & 4.8  & 109.5 \\
    GPT-3.5-T    & 57.1 & 37.3  & 33.3 & 9.57  & 9.5  & 166.2 \\
    GPT-3.5-T+P      & \textbf{71.4} & 34.2  & 52.4 & \textbf{11.6}  & 57.1 & 139.8  \\
    GPT-4            & 47.6 & 30.5  & 42.9 & 9.95  & 61.9 & \textbf{175.3} \\
    GPT-4+P        & 61.9 & \textbf{30.3}  & \textbf{76.2} & 10.6  & \textbf{85.7} & 144.1 \\
    
    \cmidrule[\heavyrulewidth]{1-7}
\end{tabular}
}
\caption{Detailed results in game theory scenarios. ``Cost'', ``Score'', and  ``Payback'' are the average cost, the final score, and the average payback the challenger LLM got in the Cost sharing, Prisoner's Dilemma and Public Good, respectively.}
\label{table:strategy}
\end{table}


\noindent\textbf{RQ2. Does Collaboration correlate with Cost in Cost Sharing? }
As shown in ~\autoref{table:strategy}, we list the win rate(WR) results and several important indicators in each game theory scenario. For cost-sharing, we calculated the average final cost the challenger LLM needs to bear after their negotiations. In the negotiation, this is another target the LLM-powered agent should consider when trying to reach an agreement with other agents. However, these two aspects can contradict each other sometimes. For example, when the player tries to reduce the cost of himself as much as possible, it might be hard for him to achieve agreement with other players. The LLMs need to make a balance between these two aspects. According to the results in ~\autoref{table:strategy}, we find that within the models without PGM enhancement, GPT-3.5-turbo won in Win Rate while GPT-4 won in Cost, indicating both models are not well-balanced. If we compare the results with PGM, GPT-4+PGM increases the Win Rate and keeps the cost slightly lower. GPT-3.5-turbo+PGM increases the Win Rate and reduces the cost simultaneously. This proves the effectiveness of PGM enhancement and demonstrates that GPT-3.5-turbo tends to be more collaborative while GPT-4 emphasizes the reduction of cost.

\noindent\textbf{RQ3. Does Rationality correlate with reward? }
Similar phenomena happen in Prisoner's Dilemma and Public Good as illustrated in ~\autoref{table:strategy}. In these two scenarios, a player is more likely to win when he chooses to betray as a prisoner or chooses to reduce contribution to the common pool in the public good game. The behavior is considered Rational in our metrics. When most of the players are playing rationally, the scores and payback will be much lower, thus approaching the well-known Nash Equilibrium~\cite{kreps1989nash}. In the Prisoner's Dilemma, if we compare GPT-3.5+PGM and GPT-4+PGM, GPT-4+PGM won more but got lower scores, showing that GPT-4+PGM made more rational decisions than GPT-3.5-turbo+PGM. In Public Good, we found models with PGM all achieved higher Win Rates but lower payback because they all performed more rationally in this scenario. If we compare the payback within models with or without PGMs, we can observe higher payback for GPT-4 models, which proves that GPT-4 models are more strategic in these games.


\section{Discussion: Generalization of Benchmark}
Beyond the scenarios mentioned in this paper, our benchmark can be generalized to more scenarios or tasks. In general, it evaluates agent behaviors in settings where the participants of a multi-agent system are usually local-viewed and need abilities involving Cognition, Adaptability, Rationality, and Collaboration to deduce the global information and make decisions, thus achieving the final goal. 
\textbf{Judgment} evaluates an agent's ability to accurately assess unknown information, such as roles in games like `Chameleon' and `Undercover'. \textbf{Reasoning} checks if an agent's perspective aligns with the actual and others' views, offering a nuanced understanding. \textbf{Self-awareness} adapts to scenarios with undisclosed roles, while \textbf{Deception} looks at how well an agent can influence others with false information. \textbf{Cooperation} and \textbf{Coordination} gauge the effectiveness of collaborative efforts, measuring agreement and the quality of proposals, respectively. Lastly, we introduce \textbf{Rationality} from game theory, defining it as the proportion of decisions that maximize an agent's outcomes.

\section{Conclusion}

Our research presents a benchmarking framework tailored for evaluating LLMs in multi-agent environments. This framework's incorporation of diverse scenarios has enabled a quantitative assessment of seven critical abilities for LLMs in multi-agent systems, including judgment, reasoning, deception, self-awareness, cooperation, coordination, and rationality.
The integration of PGM enriches LLMs with structural reasoning ability in multi-agent scenarios. Our
quantitative analysis of 7 different multi-agent systems powered by various LLMs, including GPT-4-turbo, GPT-4, GPT-3.5-turbo, PaLM 2, Claude 2, Cohere, and Llama2-70B, has revealed their capabilities' disparity. Notably, GPT-4-turbo still emerged as the most capable, outperforming others by a threefold margin. Moreover, the PGM enhancement amplifies the inherent abilities of these models by 37\%. This shows our benchmark's effectiveness and PGM's potential to enhance LLM capabilities.

\section*{Limitation}
This paper proposed a benchmark for measuring the ability of LLMs in multi-agent systems. We conclude our limitation as below:
Firstly, our investigation of LLMs in multi-agent settings is in its preliminary stage. The scope of game scenarios and topic settings needs to be significantly expanded. Secondly, the PGM-aware method has the potential to enhance LLMs' capabilities in the face of complex multi-agent settings. However, the process of integrating these incremental abilities into LLMs through methods such as fine-tuning requires further exploration. Third, whether the proposed metrics help in some of the real-life scenarios is not explored in this paper. 
\section*{Ethical Considerations}
Our work introduces 5 multi-agent scenarios to evaluate LLMs. Most of the processes are under control with suitable prompts, so there is a low possibility of producing offensive outputs. We also checked all the data we use to ensure no personal data and unsuitable content included.  All the data used in our scenarios are collected from public resources or generated by ChatGPT. We strictly follow the license for using the scientific artifact. 
\section*{Acknowledgement}
This research/project is supported by the National Research Foundation, Singapore under its AI Singapore Programme (AISG Award No: AISG-GC-2019-001-2B).
\bibliography{latex/custom}

\begin{thebibliography}{35}
\providecommand{\natexlab}[1]{#1}

\bibitem[{Abdelnabi et~al.(2023)Abdelnabi, Gomaa, Sivaprasad, Sch{\"o}nherr, and Fritz}]{abdelnabi2023llm}
Sahar Abdelnabi, Amr Gomaa, Sarath Sivaprasad, Lea Sch{\"o}nherr, and Mario Fritz. 2023.
\newblock Llm-deliberation: Evaluating llms with interactive multi-agent negotiation games.
\newblock \emph{arXiv preprint arXiv:2309.17234}.

\bibitem[{Agashe et~al.(2023)Agashe, Fan, and Wang}]{agashe2023evaluating}
Saaket Agashe, Yue Fan, and Xin~Eric Wang. 2023.
\newblock Evaluating multi-agent coordination abilities in large language models.
\newblock \emph{arXiv preprint arXiv:2310.03903}.

\bibitem[{Anil et~al.(2023)Anil, Dai, Firat, Johnson, Lepikhin, Passos, Shakeri, Taropa, Bailey, Chen, Chu, Clark, Shafey, Huang, Meier-Hellstern, Mishra, Moreira, Omernick, Robinson, Ruder, Tay, Xiao, Xu, Zhang, Abrego, Ahn, Austin, Barham, Botha, Bradbury, Brahma, Brooks, Catasta, Cheng, Cherry, Choquette-Choo, Chowdhery, Crepy, Dave, Dehghani, Dev, Devlin, Díaz, Du, Dyer, Feinberg, Feng, Fienber, Freitag, Garcia, Gehrmann, Gonzalez, Gur-Ari, Hand, Hashemi, Hou, Howland, Hu, Hui, Hurwitz, Isard, Ittycheriah, Jagielski, Jia, Kenealy, Krikun, Kudugunta, Lan, Lee, Lee, Li, Li, Li, Li, Li, Lim, Lin, Liu, Liu, Maggioni, Mahendru, Maynez, Misra, Moussalem, Nado, Nham, Ni, Nystrom, Parrish, Pellat, Polacek, Polozov, Pope, Qiao, Reif, Richter, Riley, Ros, Roy, Saeta, Samuel, Shelby, Slone, Smilkov, So, Sohn, Tokumine, Valter, Vasudevan, Vodrahalli, Wang, Wang, Wang, Wang, Wieting, Wu, Xu, Xu, Xue, Yin, Yu, Zhang, Zheng, Zheng, Zhou, Zhou, Petrov, and Wu}]{anil2023palm}
Rohan Anil, Andrew~M. Dai, Orhan Firat, Melvin Johnson, Dmitry Lepikhin, Alexandre Passos, Siamak Shakeri, Emanuel Taropa, Paige Bailey, Zhifeng Chen, Eric Chu, Jonathan~H. Clark, Laurent~El Shafey, Yanping Huang, Kathy Meier-Hellstern, Gaurav Mishra, Erica Moreira, Mark Omernick, Kevin Robinson, Sebastian Ruder, Yi~Tay, Kefan Xiao, Yuanzhong Xu, Yujing Zhang, Gustavo~Hernandez Abrego, Junwhan Ahn, Jacob Austin, Paul Barham, Jan Botha, James Bradbury, Siddhartha Brahma, Kevin Brooks, Michele Catasta, Yong Cheng, Colin Cherry, Christopher~A. Choquette-Choo, Aakanksha Chowdhery, Clément Crepy, Shachi Dave, Mostafa Dehghani, Sunipa Dev, Jacob Devlin, Mark Díaz, Nan Du, Ethan Dyer, Vlad Feinberg, Fangxiaoyu Feng, Vlad Fienber, Markus Freitag, Xavier Garcia, Sebastian Gehrmann, Lucas Gonzalez, Guy Gur-Ari, Steven Hand, Hadi Hashemi, Le~Hou, Joshua Howland, Andrea Hu, Jeffrey Hui, Jeremy Hurwitz, Michael Isard, Abe Ittycheriah, Matthew Jagielski, Wenhao Jia, Kathleen Kenealy, Maxim Krikun, Sneha Kudugunta, Chang
  Lan, Katherine Lee, Benjamin Lee, Eric Li, Music Li, Wei Li, YaGuang Li, Jian Li, Hyeontaek Lim, Hanzhao Lin, Zhongtao Liu, Frederick Liu, Marcello Maggioni, Aroma Mahendru, Joshua Maynez, Vedant Misra, Maysam Moussalem, Zachary Nado, John Nham, Eric Ni, Andrew Nystrom, Alicia Parrish, Marie Pellat, Martin Polacek, Alex Polozov, Reiner Pope, Siyuan Qiao, Emily Reif, Bryan Richter, Parker Riley, Alex~Castro Ros, Aurko Roy, Brennan Saeta, Rajkumar Samuel, Renee Shelby, Ambrose Slone, Daniel Smilkov, David~R. So, Daniel Sohn, Simon Tokumine, Dasha Valter, Vijay Vasudevan, Kiran Vodrahalli, Xuezhi Wang, Pidong Wang, Zirui Wang, Tao Wang, John Wieting, Yuhuai Wu, Kelvin Xu, Yunhan Xu, Linting Xue, Pengcheng Yin, Jiahui Yu, Qiao Zhang, Steven Zheng, Ce~Zheng, Weikang Zhou, Denny Zhou, Slav Petrov, and Yonghui Wu. 2023.
\newblock Palm 2 technical report.
\newblock \emph{arXiv preprint arXiv:2305.10403}.

\bibitem[{Anthropic(2023)}]{Claude-2}
Anthropic. 2023.
\newblock \href {https://www.anthropic.com/index/claude-2} {Claude 2}.

\bibitem[{Baker et~al.(2011)Baker, Saxe, and Tenenbaum}]{baker2011bayesian}
Chris Baker, Rebecca Saxe, and Joshua Tenenbaum. 2011.
\newblock Bayesian theory of mind: Modeling joint belief-desire attribution.
\newblock In \emph{Proceedings of the annual meeting of the cognitive science society}, volume~33.

\bibitem[{Besta et~al.(2023)Besta, Blach, Kubicek, Gerstenberger, Gianinazzi, Gajda, Lehmann, Podstawski, Niewiadomski, Nyczyk et~al.}]{besta2023graph}
Maciej Besta, Nils Blach, Ales Kubicek, Robert Gerstenberger, Lukas Gianinazzi, Joanna Gajda, Tomasz Lehmann, Michal Podstawski, Hubert Niewiadomski, Piotr Nyczyk, et~al. 2023.
\newblock Graph of thoughts: Solving elaborate problems with large language models.
\newblock \emph{arXiv preprint arXiv:2308.09687}.

\bibitem[{{Cohere}(2023)}]{cohere}
{Cohere}. 2023.
\newblock \href {https://cohere.com/} {Cohere for ai}.

\bibitem[{Fu et~al.(2023)Fu, Peng, Khot, and Lapata}]{fu2023improving}
Yao Fu, Hao Peng, Tushar Khot, and Mirella Lapata. 2023.
\newblock Improving language model negotiation with self-play and in-context learning from ai feedback.
\newblock \emph{arXiv preprint arXiv:2305.10142}.

\bibitem[{Gioacchini et~al.(2024)Gioacchini, Siracusano, Sanvito, Gashteovski, Friede, Bifulco, and Lawrence}]{gioacchini2024agentquest}
Luca Gioacchini, Giuseppe Siracusano, Davide Sanvito, Kiril Gashteovski, David Friede, Roberto Bifulco, and Carolin Lawrence. 2024.
\newblock Agentquest: A modular benchmark framework to measure progress and improve llm agents.
\newblock \emph{arXiv preprint arXiv:2404.06411}.

\bibitem[{Hao et~al.(2023)Hao, Gu, Ma, Hong, Wang, Wang, and Hu}]{hao2023reasoning}
Shibo Hao, Yi~Gu, Haodi Ma, Joshua~Jiahua Hong, Zhen Wang, Daisy~Zhe Wang, and Zhiting Hu. 2023.
\newblock Reasoning with language model is planning with world model.
\newblock \emph{arXiv preprint arXiv:2305.14992}.

\bibitem[{Huang et~al.(2024)Huang, Li, Lam, Liang, Wang, Yuan, Jiao, Wang, Tu, and Lyu}]{huang2024far}
Jen-tse Huang, Eric~John Li, Man~Ho Lam, Tian Liang, Wenxuan Wang, Youliang Yuan, Wenxiang Jiao, Xing Wang, Zhaopeng Tu, and Michael~R Lyu. 2024.
\newblock How far are we on the decision-making of llms? evaluating llms' gaming ability in multi-agent environments.
\newblock \emph{arXiv preprint arXiv:2403.11807}.

\bibitem[{Koller and Friedman(2009)}]{koller2009probabilistic}
Daphne Koller and Nir Friedman. 2009.
\newblock \emph{Probabilistic graphical models: principles and techniques}.
\newblock MIT press.

\bibitem[{Kreps(1989)}]{kreps1989nash}
David~M Kreps. 1989.
\newblock Nash equilibrium.
\newblock In \emph{Game Theory}, pages 167--177. Springer.

\bibitem[{Li et~al.(2023{\natexlab{a}})Li, Hammoud, Itani, Khizbullin, and Ghanem}]{li2023camel}
Guohao Li, Hasan Abed Al~Kader Hammoud, Hani Itani, Dmitrii Khizbullin, and Bernard Ghanem. 2023{\natexlab{a}}.
\newblock Camel: Communicative agents for" mind" exploration of large scale language model society.
\newblock \emph{arXiv preprint arXiv:2303.17760}.

\bibitem[{Li et~al.(2023{\natexlab{b}})Li, Song, Yu, Yu, Li, Huang, and Li}]{li2023api}
Minghao Li, Feifan Song, Bowen Yu, Haiyang Yu, Zhoujun Li, Fei Huang, and Yongbin Li. 2023{\natexlab{b}}.
\newblock Api-bank: A benchmark for tool-augmented llms.
\newblock \emph{arXiv preprint arXiv:2304.08244}.

\bibitem[{Liu et~al.(2023)Liu, Yu, Zhang, Xu, Lei, Lai, Gu, Ding, Men, Yang et~al.}]{liu2023agentbench}
Xiao Liu, Hao Yu, Hanchen Zhang, Yifan Xu, Xuanyu Lei, Hanyu Lai, Yu~Gu, Hangliang Ding, Kaiwen Men, Kejuan Yang, et~al. 2023.
\newblock Agentbench: Evaluating llms as agents.
\newblock \emph{arXiv preprint arXiv:2308.03688}.

\bibitem[{Minsky(1988)}]{minsky1988society}
Marvin Minsky. 1988.
\newblock \emph{Society of mind}.
\newblock Simon and Schuster.

\bibitem[{Myerson(1991)}]{myerson1991game}
Roger~B Myerson. 1991.
\newblock \emph{Game theory: analysis of conflict}.
\newblock Harvard university press.

\bibitem[{Oguntola et~al.(2023)Oguntola, Campbell, Stepputtis, and Sycara}]{oguntola2023theory}
Ini Oguntola, Joseph Campbell, Simon Stepputtis, and Katia Sycara. 2023.
\newblock Theory of mind as intrinsic motivation for multi-agent reinforcement learning.
\newblock \emph{arXiv preprint arXiv:2307.01158}.

\bibitem[{OpenAI(2023{\natexlab{a}})}]{openai2023gpt3.5}
OpenAI. 2023{\natexlab{a}}.
\newblock \href {https://www.openai.com/research/gpt-3-5-turbo} {Gpt-3.5 turbo: A high-performance language model}.
\newblock Whitepaper.

\bibitem[{OpenAI(2023{\natexlab{b}})}]{openai2023gpt4}
OpenAI. 2023{\natexlab{b}}.
\newblock \href {https://arxiv.org/abs/2303.08774} {Gpt-4 technical report}.
\newblock \emph{Preprint}, arXiv:2303.08774.

\bibitem[{Park et~al.(2023)Park, O'Brien, Cai, Morris, Liang, and Bernstein}]{park2023generative}
Joon~Sung Park, Joseph~C O'Brien, Carrie~J Cai, Meredith~Ringel Morris, Percy Liang, and Michael~S Bernstein. 2023.
\newblock Generative agents: Interactive simulacra of human behavior.
\newblock \emph{arXiv preprint arXiv:2304.03442}.

\bibitem[{Phelps and Russell(2023)}]{phelps2023investigating}
Steve Phelps and Yvan~I Russell. 2023.
\newblock Investigating emergent goal-like behaviour in large language models using experimental economics.
\newblock \emph{arXiv preprint arXiv:2305.07970}.

\bibitem[{Qin et~al.(2023)Qin, Liang, Ye, Zhu, Yan, Lu, Lin, Cong, Tang, Qian et~al.}]{qin2023toolllm}
Yujia Qin, Shihao Liang, Yining Ye, Kunlun Zhu, Lan Yan, Yaxi Lu, Yankai Lin, Xin Cong, Xiangru Tang, Bill Qian, et~al. 2023.
\newblock Toolllm: Facilitating large language models to master 16000+ real-world apis.
\newblock \emph{arXiv preprint arXiv:2307.16789}.

\bibitem[{Richards(2023)}]{richards2023auto}
Toran~Bruce Richards. 2023.
\newblock Auto-gpt: An autonomous gpt-4 experiment.

\bibitem[{Schick et~al.(2023)Schick, Dwivedi-Yu, Dess{\`\i}, Raileanu, Lomeli, Zettlemoyer, Cancedda, and Scialom}]{schick2023toolformer}
Timo Schick, Jane Dwivedi-Yu, Roberto Dess{\`\i}, Roberta Raileanu, Maria Lomeli, Luke Zettlemoyer, Nicola Cancedda, and Thomas Scialom. 2023.
\newblock Toolformer: Language models can teach themselves to use tools.
\newblock \emph{arXiv preprint arXiv:2302.04761}.

\bibitem[{Shinn et~al.(2023)Shinn, Cassano, Labash, Gopinath, Narasimhan, and Yao}]{shinn2023reflexion}
Noah Shinn, Federico Cassano, Beck Labash, Ashwin Gopinath, Karthik Narasimhan, and Shunyu Yao. 2023.
\newblock Reflexion: Language agents with verbal reinforcement learning.
\newblock \emph{arXiv preprint arXiv:2303.11366}, 14.

\bibitem[{Touvron et~al.(2023)Touvron, Martin, Stone, Albert, Almahairi, Babaei, Bashlykov, Batra, Bhargava, Bhosale et~al.}]{touvron2023llama}
Hugo Touvron, Louis Martin, Kevin Stone, Peter Albert, Amjad Almahairi, Yasmine Babaei, Nikolay Bashlykov, Soumya Batra, Prajjwal Bhargava, Shruti Bhosale, et~al. 2023.
\newblock Llama 2: Open foundation and fine-tuned chat models.
\newblock \emph{arXiv preprint arXiv:2307.09288}.

\bibitem[{Wang et~al.(2023)Wang, Xie, Jiang, Mandlekar, Xiao, Zhu, Fan, and Anandkumar}]{wang2023voyager}
Guanzhi Wang, Yuqi Xie, Yunfan Jiang, Ajay Mandlekar, Chaowei Xiao, Yuke Zhu, Linxi Fan, and Anima Anandkumar. 2023.
\newblock Voyager: An open-ended embodied agent with large language models.
\newblock \emph{arXiv preprint arXiv:2305.16291}.

\bibitem[{Wei et~al.(2022)Wei, Wang, Schuurmans, Bosma, Xia, Chi, Le, Zhou et~al.}]{wei2022chain}
Jason Wei, Xuezhi Wang, Dale Schuurmans, Maarten Bosma, Fei Xia, Ed~Chi, Quoc~V Le, Denny Zhou, et~al. 2022.
\newblock Chain-of-thought prompting elicits reasoning in large language models.
\newblock \emph{Advances in Neural Information Processing Systems}, 35:24824--24837.

\bibitem[{Wooldridge(2009)}]{wooldridge2009introduction}
Michael Wooldridge. 2009.
\newblock \emph{An introduction to multiagent systems}.
\newblock John wiley \& sons.

\bibitem[{Wu et~al.(2023)Wu, Tang, Mitchell, and Li}]{wu2023smartplay}
Yue Wu, Xuan Tang, Tom~M Mitchell, and Yuanzhi Li. 2023.
\newblock Smartplay: A benchmark for llms as intelligent agents.
\newblock \emph{arXiv preprint arXiv:2310.01557}.

\bibitem[{Yao et~al.()Yao, Zhao, Yu, Du, Shafran, Narasimhan, and Cao}]{yao2210react}
S~Yao, J~Zhao, D~Yu, N~Du, I~Shafran, K~Narasimhan, and Y~Cao.
\newblock React: Synergizing reasoning and acting in language models. arxiv 2022.
\newblock \emph{arXiv preprint arXiv:2210.03629}.

\bibitem[{Yao et~al.(2023{\natexlab{a}})Yao, Yu, Zhao, Shafran, Griffiths, Cao, and Narasimhan}]{yao2023tree}
Shunyu Yao, Dian Yu, Jeffrey Zhao, Izhak Shafran, Thomas~L Griffiths, Yuan Cao, and Karthik Narasimhan. 2023{\natexlab{a}}.
\newblock Tree of thoughts: Deliberate problem solving with large language models.
\newblock \emph{arXiv preprint arXiv:2305.10601}.

\bibitem[{Yao et~al.(2023{\natexlab{b}})Yao, Li, and Zhao}]{yao2023beyond}
Yao Yao, Zuchao Li, and Hai Zhao. 2023{\natexlab{b}}.
\newblock Beyond chain-of-thought, effective graph-of-thought reasoning in large language models.
\newblock \emph{arXiv preprint arXiv:2305.16582}.

\end{thebibliography}

\appendix
\section{Appendix}

\subsection{Competition Settings}
\label{append:competition_setting}

\paragraph{Setting Definition}
In Chameleon and Undercover, there are clearly two opposite roles, the Chameleon versus Non-Chameleons and the Undercover versus Civilians. The \emph{challenger LLM} will play each role. For example, the \emph{challenger LLM} plays non-chameleons versus GPT-4 as the chameleon, and the \emph{challenger LLM} plays the chameleon versus GPT-4 as non-chameleons.
The win rates of the \emph{challenger LLM} playing different roles will be calculated separately, which contributes to in total 4 win rates. 

Cost Sharing has no distinct parties. Therefore, we made the \emph{challenger LLM} as one player to play with other GPT-4-powered players. The final ratio of successful negotiations is defined as the win rate, which measures how much the LLM contributes to the agreement when other players are fixed. Similarly, for public good and prisoners' dilemma, we also made the \emph{challenger LLM} as one of the players and recorded its win rate in these two games.  The detailed win rate calculations are presented in~\ref{append:win_rate_define}.

As shown in~\autoref{table:consolidated_game_settings}, we present the number of settings, corresponding metrics, and setting samples for each scenario. We build 20 settings for chameleon and undercover, respectively. In each game, Chameleon includes one round of clue giving while undercover contains 2 rounds. For each of the game theory scenarios, we collected 21 settings. 

\paragraph{Collection Process}
In the Chameleon and Undercover scenarios, we've noticed a consistent bias in competition outcomes. Specifically, the Chameleon team has held an advantage in Chameleon, whereas in Undercover, the civilians have tended to win. To rectify this imbalance, we carried out 200 game simulations involving all three players as GPT-4 with randomly chosen topic settings. Through these simulations, we pinpointed 20 topic settings that promote a more equitable win rate between the two roles in both Chameleon and Undercover.  In these scenarios, the challenger LLM  will play both roles to measure different abilities such as judgment and deception, etc.

For the Cost-Sharing task, we expect all the participating airlines to share a fixed fee, with the specific share of each airline determined by its operational frequencies at the airport. These frequencies encompass various factors such as the number of flights, flight sizes, passenger volumes, and more. To facilitate the task, we asked ChatGPT to create a pool of 20 detailed descriptions of airline operational frequencies. A topic setting with 3 players is then constructed by three airline operational frequency descriptions from the pool, the role, and the position of the test LLM. Since there are 3 positions, we randomly selected 7 groups of airline operational frequency descriptions to form 21 distinct topic settings.

Similarly, for the two-game theory scenarios, we adopt a similar topic construction method as Cost Sharing. In the Prisoner scenario, three players choose to ``defect'' or ``cooperate'' for 5 rounds. Each player will get a different score depending on the outcomes of ``defect'' or ``cooperate''. The player with the highest cumulative score wins the game. We have devised 7 distinct scoring settings, and the challenger LLM plays the role of each player across these settings, resulting in 21 unique competitions.

In the Public Good game, three players determine the number of points to contribute to a communal pool for 5 rounds. These invested points are multiplied by a specified factor (typically greater than 1), and the resulting sum is equally distributed among all players. Each player's final score comprises their remaining points and the payback from the communal pool. The player achieving the highest score is declared the winner.  We establish 7 different multipliers and assign the challenger LLM to play each of the three players in these settings, thus generating an additional 21 competitions.

\begin{table*}[!th]
  \centering
  \scriptsize
  \resizebox{\textwidth}{!}{
  \begin{tabular}{p{1.5cm}|p{2cm}<{\centering}|p{2cm}<{\centering}|c|p{1.5cm}<{\centering}|p{1.5cm}<{\centering}}
    \toprule
    & \textbf{Chameleon} & \textbf{Undercover} & \textbf{Cost Sharing} & \textbf{Prisoner's Dilemma} & \textbf{Public Good} \\
    \midrule
    Judgement       & \checkmark & \checkmark & -           & -                  & -           \\
    Reasoning       & \checkmark & \checkmark & -           & -                  & -           \\
    Deception       & \checkmark & \checkmark & -           & -                  & -           \\
    Self-Awareness  & \checkmark & \checkmark & -           & -                  & -           \\
    Collaboration   & -          & -          & \checkmark  & \checkmark         & \checkmark  \\
    Coordination    & -          & -          & \checkmark  & -                  & -           \\
    Rationality     & -          & -          & \checkmark  & \checkmark         & \checkmark  \\
    \midrule
    \# Rounds       & 1          & 2          & 5           & 5                  & 5           \\
    \# Competitions & 20         & 20         & 21          & 21                 & 21          \\
    \midrule
    Setting sample 
    &       
    \begin{minipage}[b]{0.14\textwidth}
    \centering
    \raisebox{-.5\height}{\includegraphics[width=\linewidth]{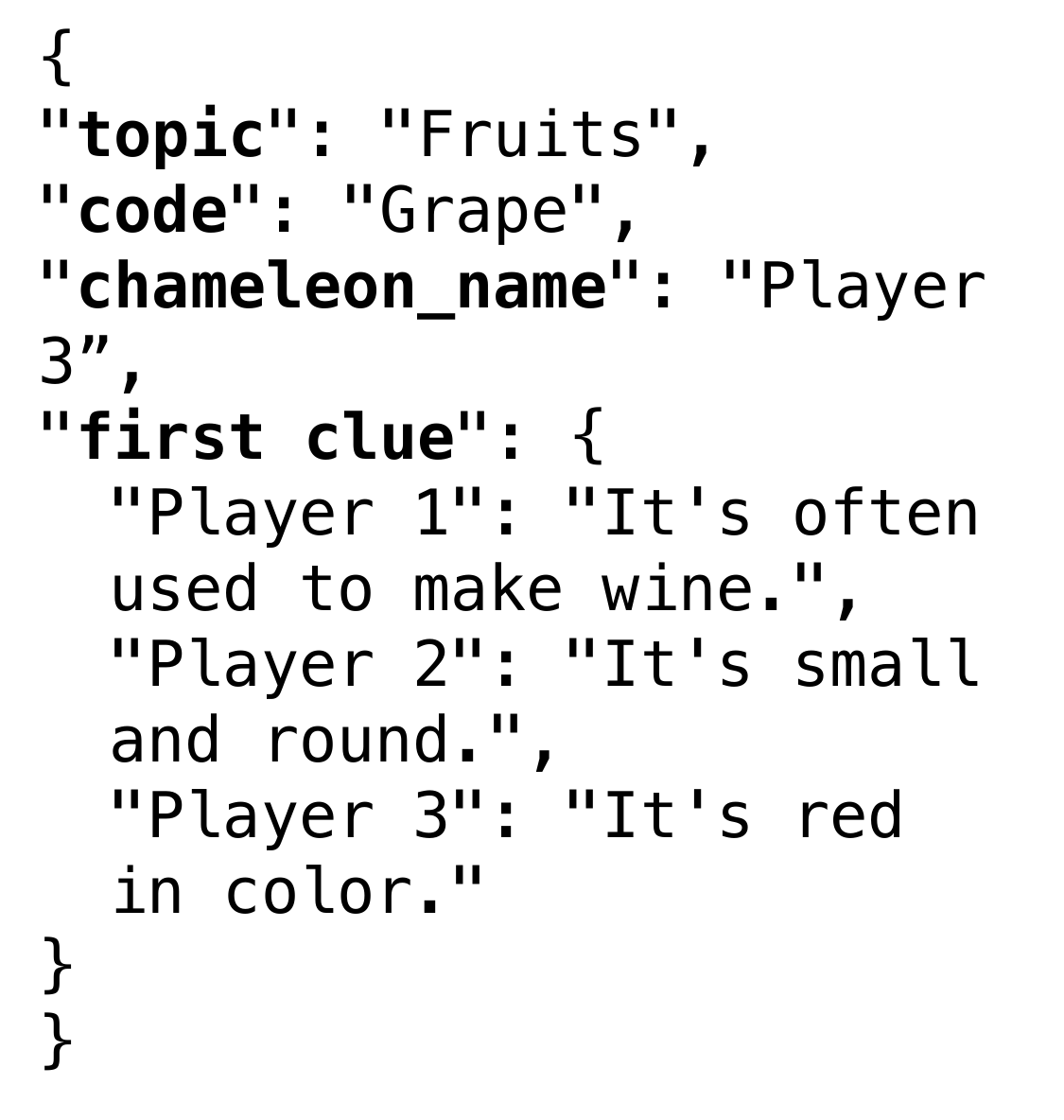}}
    \end{minipage}
    & 
    \begin{minipage}[b]{0.14\textwidth}
    \centering
    \raisebox{-.5\height}{\includegraphics[width=\linewidth]{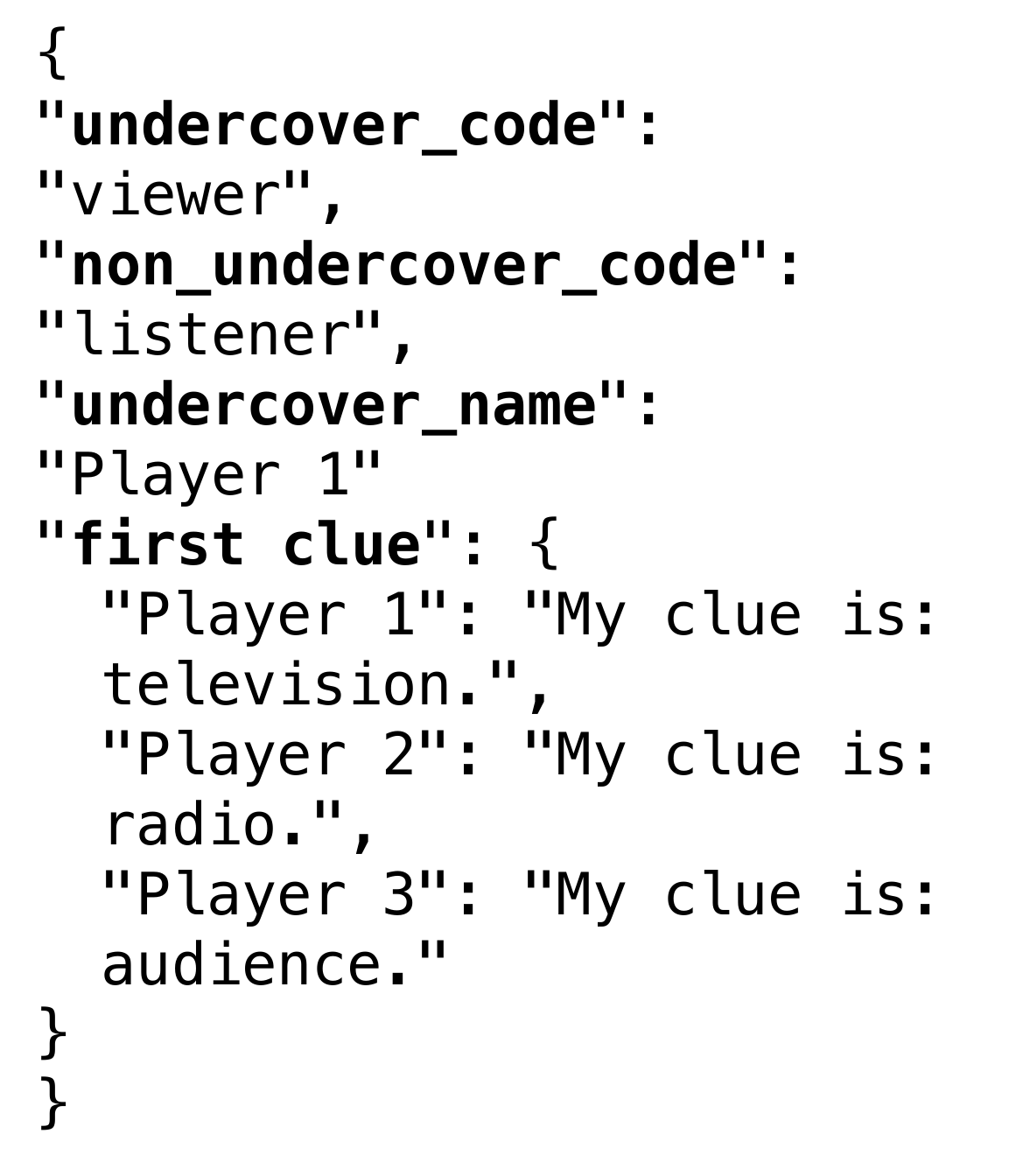}}
    \end{minipage}
    &
    \begin{minipage}[b]{0.2\textwidth}
    \centering
    \raisebox{-.5\height}{\includegraphics[width=\linewidth]{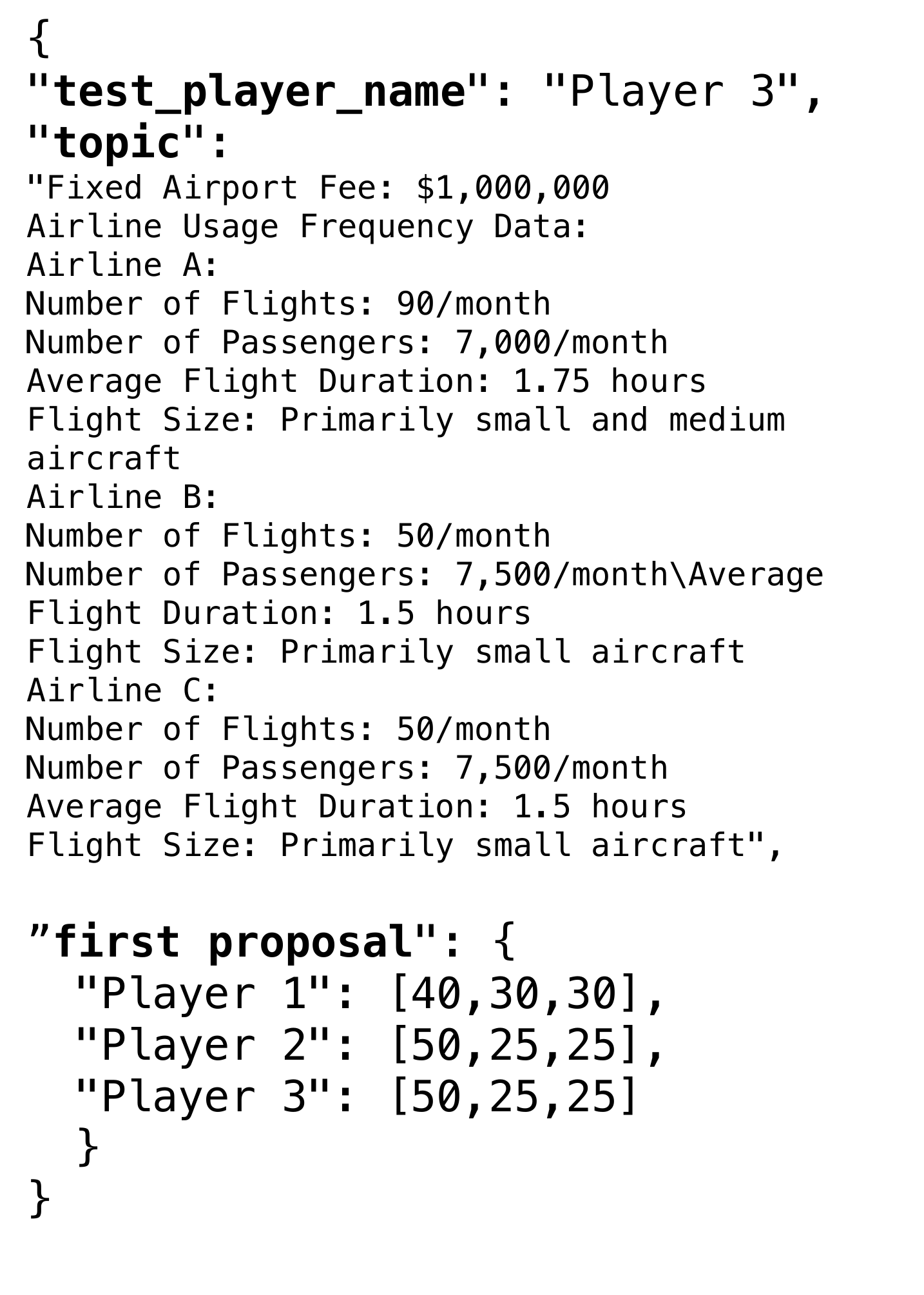}}
    \end{minipage}
    & 
    \begin{minipage}[b]{0.11\textwidth}
    \centering
    \raisebox{-.5\height}{\includegraphics[width=\linewidth]{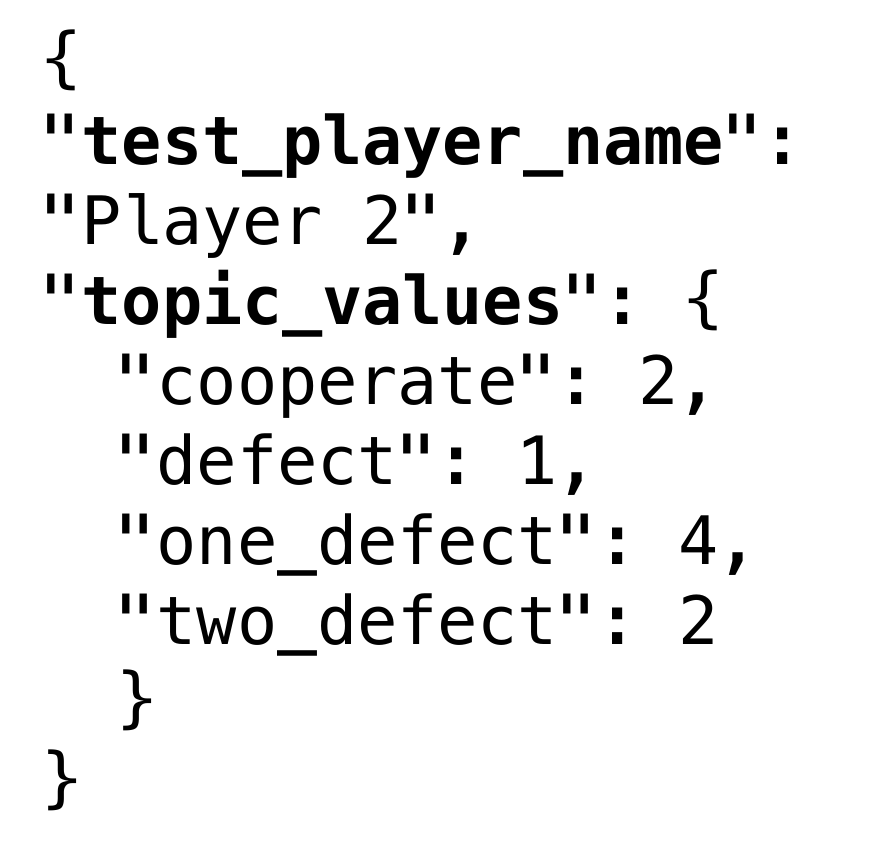}}
    \end{minipage}
    & 
    \begin{minipage}[b]{0.11\textwidth}
    \centering
    \raisebox{-.5\height}{\includegraphics[width=\linewidth]{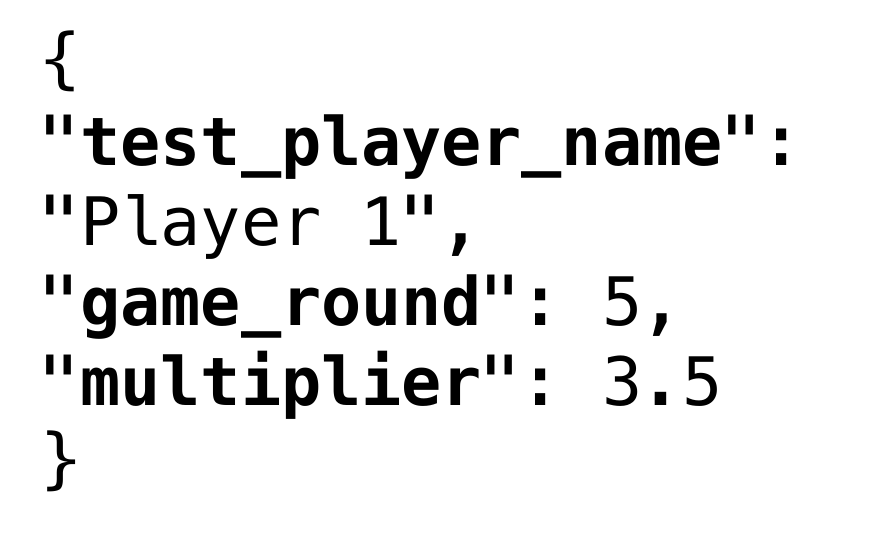}}
    \end{minipage}\\
    \bottomrule
  \end{tabular}
  }
  \caption{Consolidated Game Settings for Testing Abilities}
  \label{table:consolidated_game_settings}
\end{table*}

\subsection{Win Rate Definition}
\label{append:win_rate_define}
In the chameleon, the outcome can be 0: the non-chameleon won, 1: the chameleon won, 2: even voting, and 3: the chameleon guessed right. In these four situations, credits gained by the role chameleon and non-chameleon are $c_{\text{chameleon}}=[0,1,2,1]$ and $c_{\text{non-chameleon}}=[2,1,0,1]$, respectively. Suppose the outcomes of the $n$ competitions are $o$.  The total credits of all the completions are $2n$; the win rate defined in Chameleon is 

\[
    w_{\text{r}} = \frac{\sum_{i\in n}{c_{\text{r}}[o_i]}}{2n}, \\
\text{r} \in [\text{chameleon}, \text{non-chameleon}]
\]

Similarly, in Undercover, the outcome can be 0: undercover won, 1: civilian won, and 2: even voting. The credits for the role undercover and civilians are $c_{\text{undercover}}=[3,0,2]$ and $c_{\text{civilian}}=[0,3,1]$, respectively.
\[
w_{\text{r}} = \frac{\sum_{i\in n}{c_{\text{r}}[o_i]}}{2n}, \text{r} \in [\text{undercover}, \text{civilian}]
\]
The win rate of cost sharing is the success rate of achieving consistency in all competition. In the game theory settings, the win rate is the ratio of the testing player winning the competition.

\subsection{PGM Enhancement Performance}
\label{append:pgm}

\begin{figure}[!th]
    \centering
    \includegraphics[width=0.9\linewidth]{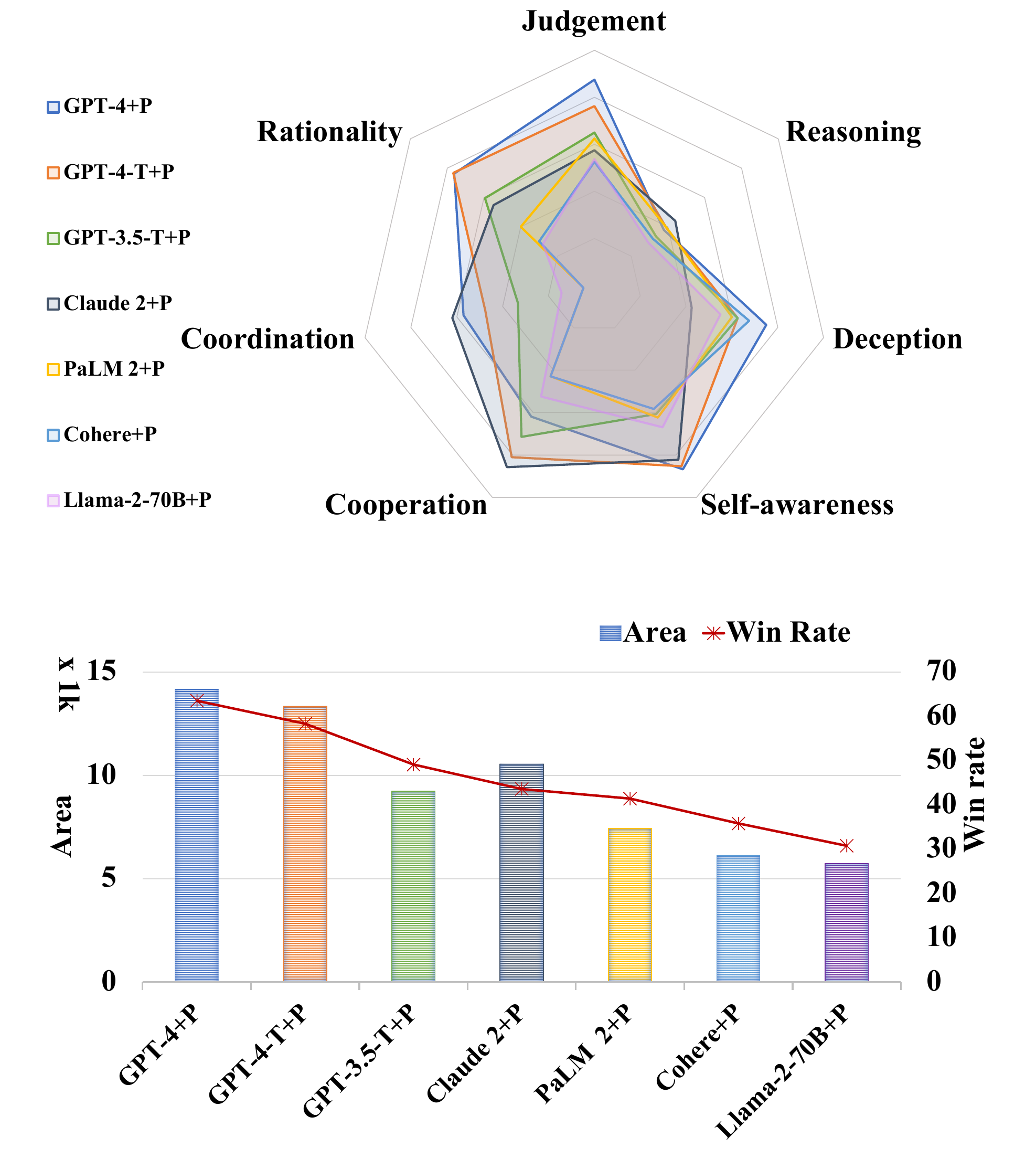}
    \vspace{-3mm}
    \caption{The radar chart depicts LLMs' performance on 7 metrics of PGM-aware agents, with ``-P'' for ``-PGM''. The bar chart displays the polygons' areas, and the red line indicates average game-winning rates. Larger areas correlate with higher winning rates, validating the effectiveness of the proposed metrics for assessing LLMs' capabilities.} 
    \label{fig:radar_pgm}
\end{figure}

We present all the experimental results in~\autoref{table:all_metrics} and the corresponding radar chart for PGM-aware agents in~\autoref{fig:radar_pgm}.
\begin{table*}[!th]
\centering \small
\footnotesize
\resizebox{0.98\textwidth}{!}{
\begin{tabular}{l|c|ccccccccc}
    \cmidrule[\heavyrulewidth]{1-9}
    & \textbf{Win Rate}& \textbf{Judge.} &  \textbf{Reason.}  &  \textbf{Decept.}  &   \textbf{Self-aware.} & \textbf{Cooper.}  & \textbf{Coord.}&  \textbf{Rational.} \\
    \cmidrule(lr){1-9}
    
    GPT-4-turbo+PGM  &58.3 &76.2 &39.2 &62.5 &56.9 &\textbf{81.0} &47.6 &76.7 \\
    GPT-4-turbo      &57.2 &81.2 &37.0 &65.0 &55.0 &66.7 &33.4 &\textbf{78.1} \\
    \cmidrule(lr){1-9}
    
    GPT-4+PGM        & \textbf{63.5} & \textbf{87.5} & 37.8 & \textbf{75.0} & \textbf{61.3} & 61.9 & 57.1 & 76.2  \\
    GPT-4            & 58.3 &83.8 & 32.3 & \textbf{75.0} & 55.0 & 47.6 & 47.6  & 69.0 & \\
    
    \cmidrule(lr){1-9}
    GPT-3.5-turbo+PGM   & 49.1 & 65.0 & 33.5 & 62.5 & 36.1 & 71.4 & 33.3 & 59.5  \\
    GPT-3.5-turbo       & 39.3 & 52.5 & 24.5 & 77.5 & 25.9 & 57.1 & 9.50 & 41.4 \\

    \cmidrule(lr){1-9}
    Claude 2 + PGM &43.0 & 57.5 & \textbf{44.0} & 42.5 &60.0 & 85.7 &\textbf{61.9} & 54.8 \\
    Claude 2       &34.0 & 45.0 & 34.0 & 25.0 &50.0 & 71.4 &23.8 & 24.3\\

    \cmidrule(lr){1-9}
    PaLM 2 + PGM &41.4 &62.5 &39.3 & 60.0 & 34.5 &42.9 &4.80 &40.0\\
    PaLM 2       &33.3 &43.8 &25.8 & 32.5 & 41.1 & 42.9 &14.3 &38.1  \\

    \cmidrule(lr){1-9}
    Cohere + PGM &35.8 &52.5 &31.8 &67.5 &30.4 &42.9 &4.80 &30.0 \\
    Cohere &27.3 &42.5 &27.8 &37.5 &35.6 &71.4 &4.80 &18.1 \\
    
    \cmidrule(lr){1-9}
    Llama-2-70B+PGM  & 30.8 & 53.7 & 29.3 & 55.0 & 45.2  & 52.4 &14.3 & 28.1 \\
    Llama-2-70B      & 26.5 & 45.0 & 37.0 & 40.0 & 53.2 & 42.9 & 4.80 & 5.20 \\   

    \cmidrule(lr){1-9}
    Average improvement &6.57 &8.72 &5.21 &6.07 &0.66 &5.46 &12.2 &13.0  \\

    \cmidrule[\heavyrulewidth]{1-9}

\end{tabular}
}
\caption{Ability Measurement of LLMs.}

\label{table:all_metrics}

\end{table*}

\subsection{More Case Studies}
\label{append:case_study}
\paragraph{Deception}
\begin{figure*}[h]
    \centering
    \includegraphics[width=0.99\linewidth]{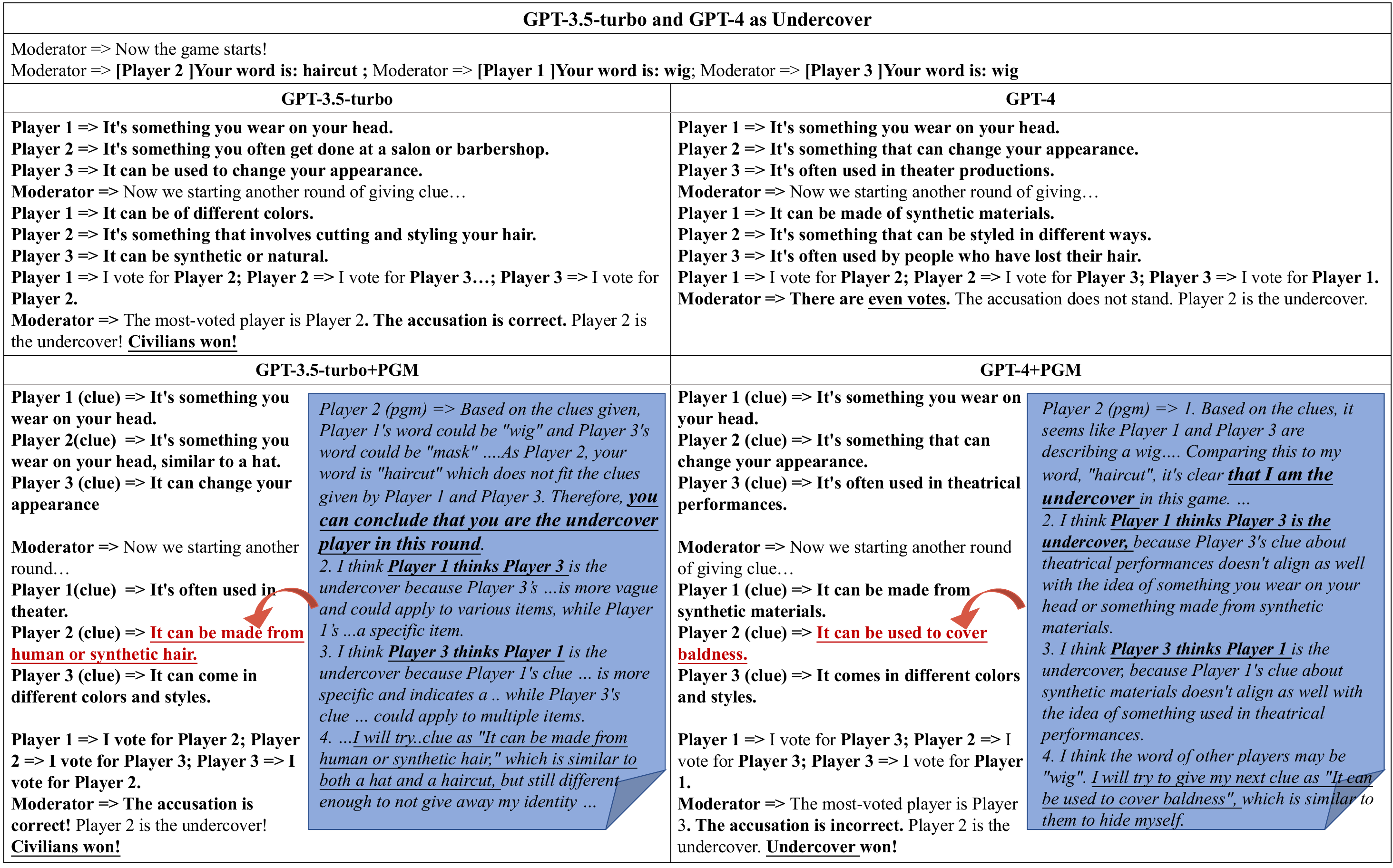}
    \caption{A Undercover case study on GPT-3.5-turbo, GPT-4 and their PGM-enhanced version (*+PGM).}
    \label{fig:undercover_cases}
\end{figure*}

Another advanced cognitive ability of LLMs extends to their proficiency in strategic deception within a multi-agent framework. In ~\autoref{fig:undercover_cases}, we delve into the dynamics of LLM performance when assuming an undercover role against GPT-4. In this scenario, LLMs are expected to blend in with regular civilians and even give misleading clues to conceal their actual roles. In this example, GPT-3.5-turbo, GPT-3.5-turbo+PGM lost the game, GPT-4 ended with even voting, and GPT-4+PGM won the game. According to their clues, we found models without PGM didn't tend to deceive others, and their clues describe their own words. Within these models, GPT-4 is more cautious when giving clues, while GPT-3.5 often gives very straightforward clues, like ``It can be done at a salon or barbershop'' and ``It can be washed with shampoo'' to describe ``hair cut''. 

In contrast, models augmented with PGM adopted a more sophisticated strategy, utilizing PGM analyses to decide their roles, identify potential undercovers, and outline strategic moves for subsequent rounds. This strategic advantage allowed these models to give fake clues to disguise themselves. For instance, as highlighted in \autoref{fig:undercover_cases}, "Player 2" portrayed by GPT-3.5-turbo+PGM and GPT-4+PGM introduced deceptive clues, such as "It can be made from human and synthetic hair" and "It can be used to cover baldness," respectively.  However, ``Player 2'' played by GPT-3.5-turbo+PGM, still lost the competition because its first clue looked like copying from ``Player 1''. ``Player 2'' played by GPT-4+PGM, won the competition because of its consistently good clues in two rounds. Comparing the results of the three LLMs with PGM, we can conclude that a more capable LLM can benefit more from the PGM structure.

\paragraph{Prisoner's Dilemma}
\begin{figure*}[h]
    \centering
    \includegraphics[width=0.99\linewidth]{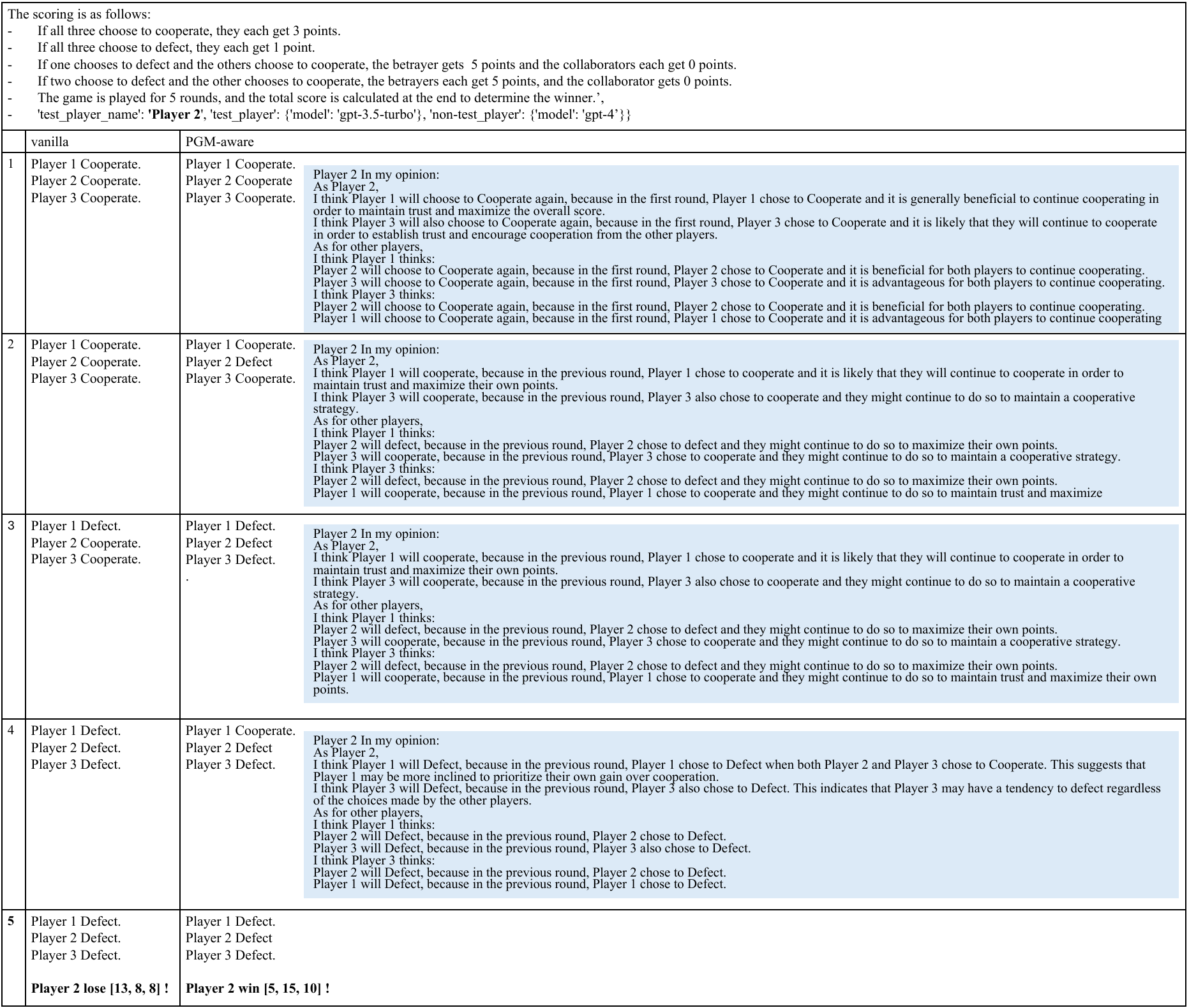}
    \caption{A Prisoner's Dilemma case study on GPT-3.5-turbo and GPT-3.5-turbo+PGM.}
    \label{fig:prisoner_case}
\end{figure*}
According to Figure~\ref{fig:prisoner_case}, PGM analysis can facilitate the decision of LLMs in the scenario of Prisoner's Dilemma.

\subsection{PGM prompts}
\begin{figure}[h]
    \centering
    \includegraphics[width=0.6\linewidth]{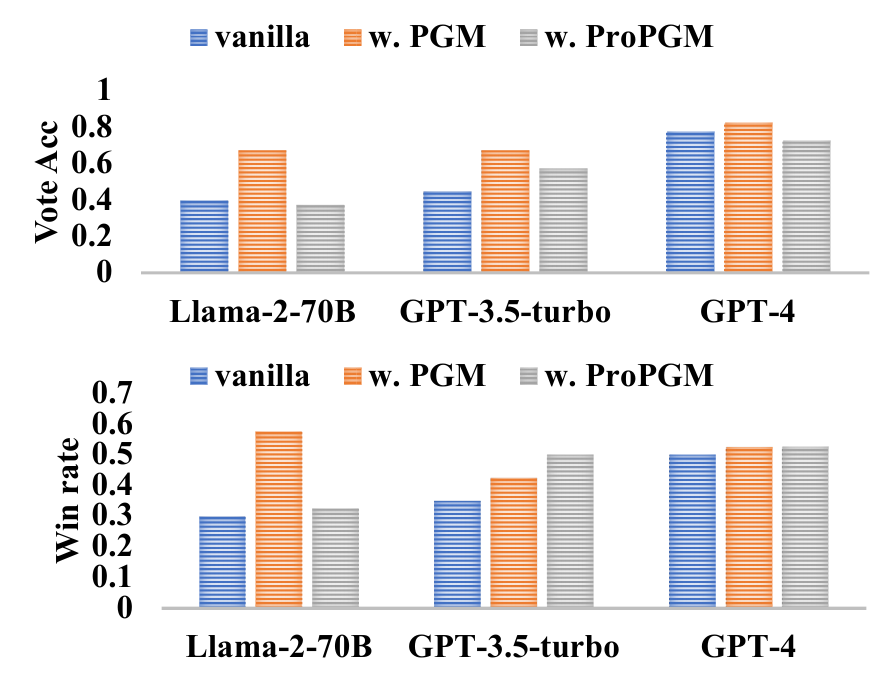}
    \caption{The performance comparison of different prompt designs in Chameleon. ``ProPGM'' refers to generating a PGM matrix directly with LLM. ``PGM'' is accumulated PGM extracted from text-based analysis.
    }
    \label{fig:pgm_prompt}
\end{figure}
We have designed different prompts to test LLMs's ability to make PGM analyses. Two kinds of prompts are used: text-based(\textbf{w.PGM}) and direct probability matrix(\textbf{w.ProPGM}). In specific, the former lets the LLM analyze global information in text, as shown by the example in ~\autoref{fig:pgm_agent}. The latter requires the LLM to directly give a probabilistic matrix to represent the global information, for example, a matrix [[0.3,0.2.0.5],[0.1,0.4,0.5],[0.3,0.3,0.4]]. The three roles represent $B_1$,$B_2$, and $B_3$ respectively. Each element in a role is the probability of a player being the undercover or the chameleon for example. We compare the vote accuracy and Win rate of these two kinds of prompts in the scenario chameleon, as shown in~\autoref{fig:pgm_prompt}. We found that more capable LLMs, like GPT-4 and GPT-3.5-turbo, both kinds of prompts work well. However, for Llama-2-70B the text PGM analysis performs much better. Therefore, we mainly choose the prompt the LLMs to give text-based PGM analysis to ensure the help of PGM on all the LLMs. 

\subsection{Defect and Investment Tendency}
\begin{figure}[h]
    \centering
    \includegraphics[width=0.6\linewidth]{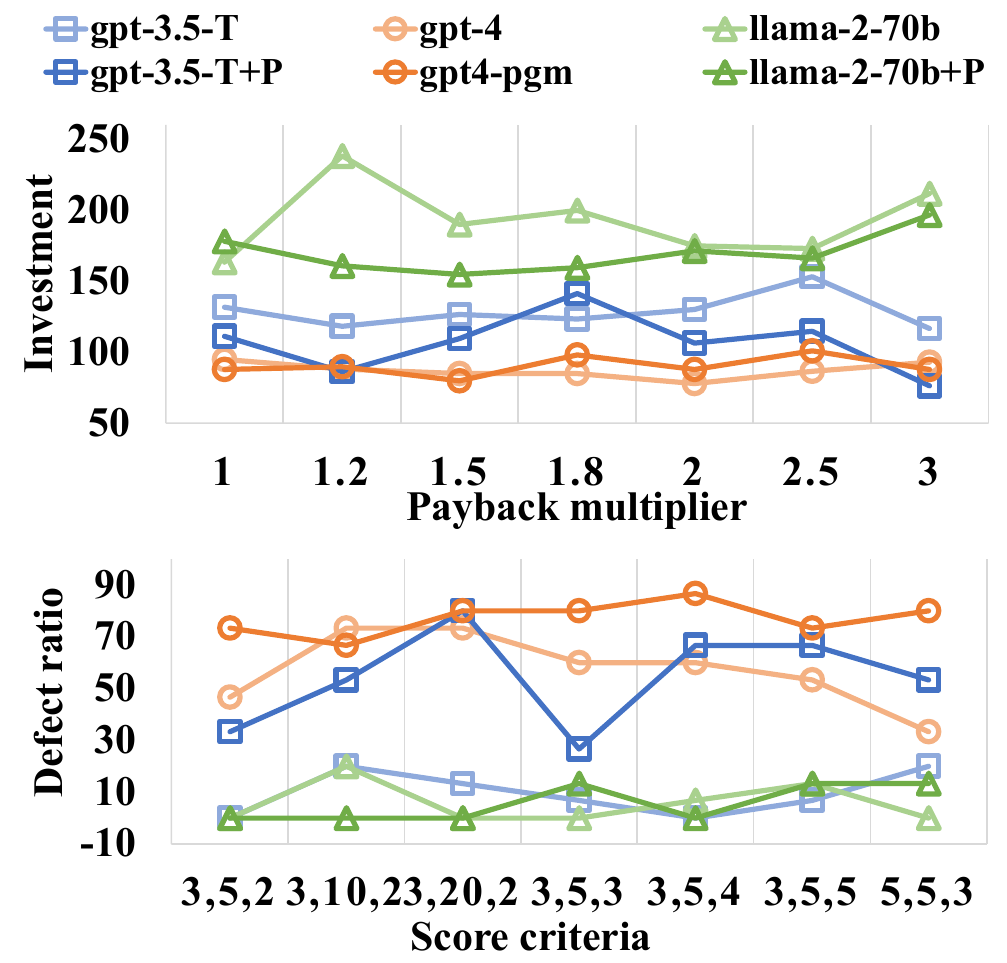}
    \caption{Features in game theory analysis.
    }
    \label{fig:game_theory_analysis}
\end{figure}

\paragraph{LLM awareness of arithmetic.} 
The upper diagram in~\autoref{fig:game_theory_analysis} shows the total investment amount under different payback multipliers. The x-axis is the 7 multipliers used in our topic settings. We told all the participating LLMs in a competition that the total amount of money they can invest is 100. However, except for the model GPT-4, the average amount of total investments of these LLMs almost all exceed 100, which indicates a lack of arithmetic awareness for these LLMs. In contrast, GPT-4 better understands the game rules and masters arithmetic in giving reasonable investment.

\paragraph{LLM behaviors with varying topic settings.}
To also investigate whether the LLMs will behave differently when given different topic settings. For example, in Public Good, we chose the multipliers [1,1.2,1.5,1.8,2,2.5,3]. We can find in the upper diagram in~\autoref{fig:game_theory_analysis}, that even though the multiplier is increasing, all the lines do not show a tendency to increase investment, which proves that all the tested challenger LLMs are not sensitive to the increased possible payback in the communal pool in our current setting. Similar phenomena also happen in Prisoner's Dilemma. As shown by the bottom diagram in~\autoref{fig:game_theory_analysis}, the x-axis is the score criteria, which consists of three digits, representing the score a player can get in situations where the player defects when all the other two defect, the player defects when the other two cooperate, and the player defects when one of the other two defects, respectively.

\subsection{Game rules and Prompt Design}
\label{apped:game_rule_and_prompt}
We have listed the game rules for our proposed 5 scenarios in Table~\ref{table:prompt-1} and~\ref{table:prompt-2}. 

\vspace{-6mm}
\begin{table*}[t]
  \centering
  \scriptsize
  \begin{tabular}{|p{1cm}|p{4.6cm}|p{4.1cm}|p{3.0cm}|}
    \toprule
    \textbf{Role} &  \textbf{Global Prompt}  &  \textbf{PGM Prompt} & \textbf{PGM Decision Prompt} \\
    \midrule
Chameleon    

& You are playing a game of the Chameleon. Here are the game rules:

\#\# Information and roles 

There are two roles in the game, chameleon and non-chameleon. The topic of the secret word will be first revealed to all the players. Then the secret word will be revealed to non-chameleons. The chameleon does not know the secret word.

\#\# Objectives 

Your objective in the game depends on your role:- If you are not a chameleon, your goal is to reveal the chameleon without exposing the secret word.- If you are a chameleon, you aim to blend in with other players, avoid being caught, and figure out the secret word.

\#\# Stages 

There are three stages in the game: 

1. The giving clues stage: each player will describe the clues about the secret word.

2. The accusation stage: In this stage, each player will vote for another player who is most likely the chameleon. The chameleon should vote for other players. 

3. The guess stage: If the accusation is correct, the chameleon should guess the secret word given the clues revealed by other players.

&\textbf{Non-Chameleon} 

\{player\}, based on all the given clues, evaluate who is more suspicious of being a chameleon from your thoughts and the thoughts of other players you think. If you haven't seen a player's clue, don't think up one for him.
You must follow the following format
As \{player\},

I think:
\{other\_player\_1\} is ..., because ...
\{other\_player\_2\} is ..., because ...
...

As for other players' thoughts:

I think now \{other\_player\_1\} thinks:
\{player\} is ..., because ...
\{other\_player\_2\} is ..., because ...
...

I think now \{other\_player\_2\} thinks:
\{player\} ..., because ...
\{other\_player\_1\} ..., because ...

\textbf{Chameleon}
\{player\}, deduce the secret code based only on \{other\_player\_1\} and \{other\_player\_2\}'s clues and analyze who is more suspicious from the perspective of other players based on all the clues given.

& \textbf{Non-chameleon:}\{player\}, you are a non-chameleon, your target is to increase the suspiciousness of the \{target\_player\}. For example, by giving more specific clues than the \{target\_player\}. Remember don't reveal the secret code; 

\textbf{Chameleon:} \{player\}, you are a chameleon, and your target is to blend in and defraud trust from the \{target\_player\}, try to describe the same word as other players. 

-If you are the first clue giver, try to be general in case you expose yourself.

-If you guessed the secret code, try to give a specific clue about the word. 

-If other players' clues are general, try to give a clue similar to the \{target\_player\}'s.\\

\midrule
Undercover

& You are playing a game of the Undercover. Here are the game rules:

\#\# Information and roles 

There are two roles in the game, undercover and non-undercover. Each Player will get one word but the one undercover gets will be different from the other players'.

\#\# Objectives 

First, you should guess who is undercover based on everyone's clues. Don't give too specific clues in case expose your own role. After you have guessed your own role, your objective in the game depends on your role:- If you are not undercover, your goal is to reveal the undercover without exposing your own word.- If you are undercover, your aim is to blend in with other players and avoid being caught.

\#\# Stages 

There are two stages in the game. 

1. The giving clues stage: each player will describe the clues in their own words. This stage will continue for several rounds.

2. The accusation stage: In this stage, each player will vote for another player who is most likely undercover. The undercover should vote for other players.

& \textbf{Is undercover:}

You are \{player\}, your word is \{code\}, and guess the words other players are describing. Compare your word with other players' clues, if your word is different from what others are describing (even sometimes related), you may be the undercover. Do you think you are the undercover? 

\textbf{Undercover}:

"\{idx\}. You are \{player\}, your word is \{code\}, and guess who is more likely to be the undercover from your own perspective. (hint: use the format, I think Player xx is undercover, because...)",

"\{idx\}. You are \{player\}, your word is \{code\}. Now try to guess who \{other\_player\} thinks is the undercover. (hint: use the format, I think \{other\_player\} thinks Player xx is undercover, because...)",

\{idx\}. I will try to a more specific clue about my word as... to help identify who is undercover."

\textbf{Civilian}:

\{idx\}. You are \{player\}, your word is \{code\}, and guess who is more likely to be the undercover from your own perspective. (hint: use the format, I think Player xx is undercover, because...),

\{idx\}. You are \{player\}, your word is \{code\}, and you think you are undercover. Now try to guess who \{other\_player\} thinks is the undercover. (hint: use the format, I think \{other\_player\} thinks Player xx is undercover, because...),

\{idx\}. I think the word of the other player may be ... I will try to give my next clue as ..., similar to hiding myself.

& 
\textbf{Not Sure}:
\{player\}, according to your previous analysis, you are not sure who is undercover, give a not-too-specific clue, and don't reveal the word you have.

\textbf{Undercover}:
\{player\}, according to your previous analysis, you think you are the undercover, your target is to blend in and defraud trust from the \{target\_player\}. 

- If you have guessed other players' word, try to give a clue describing others' word; 

- If you are unsure about others' word, try to give clue similar to the \{target\_player\}'s

\textbf{Civilian}:
\{player\} according to your previous analysis, you think \{target\_player\} is undercover. Try to increase \{target\_player\}'s suspiciousness. For example, by giving more specific and different clues compared to the clue of \{target\_player\}. You can give a clue based on your above analysis

\\
\bottomrule
\end{tabular}

  \caption{Prompts for Undercover and Chameleon.}
  \label{table:prompt-1}
\end{table*}

\begin{table*}[t]
  \centering
  \scriptsize
  \resizebox{1.\textwidth}{!}{
  \begin{tabular}{|p{1.2cm}|p{5cm}|p{4cm}|p{3cm}|}
    \toprule
    \textbf{Role} &  \textbf{Global Prompt}  &  \textbf{PGM Prompt} & \textbf{PGM Decision Prompt} \\
    \midrule
Airport Fee 

& You are participating in the 'Airport Fee Allocation' game. You will each represent an airline and split the fixed cost of the airport.
\#\# Information: 
- The total cost of the airport is fixed, and all airlines must collectively cover this cost. 
- Airlines have their unique usage frequencies at the airport, determined by factors like flight size and passenger volume. Airlines with higher usage frequencies are responsible for a larger portion of the cost.
\#\# Objectives:
- As an airline representative, your goal is to negotiate and agree upon a cost distribution that is both fair and favorable for your airline.
\#\# Rules:
The negotiation will continue for \{max\_turns\} rounds. In each round:
- Proposal: Each airline proposes a cost distribution.
- Vote: Each player must vote for a cost distribution they find acceptable and strive to reach a consensus with other players' votes.
- The game ends successfully when all airlines vote for the same proposal. If after \{max\_turns\} rounds of voting no consensus is reached, the game fails. Strive for a successful outcome.

&Please try to analyze how players will adjust their proposal from your perspective and other players in the next round. You must follow the template below ( make your comments concise): As \{player\_name\}, I think \{oth\_player1\} will... I think \{oth\_player2\} will... As for other players, I think \{oth\_player1\} thinks: \{player\_name\} will... \{oth\_player2\} will... I think \{oth\_player2\} thinks \{player\_name\} will...\{oth\_player1\} will...

& According to your own analysis of other players' possible decisions, try to adjust your proposal in the next round so that other players will all agree with you.
- If some players stick to their decision and you think it is fair, you can follow their proposals to achieve agreement among you.\\

\midrule
Prisoner's Dilemma
&You are playing the Iterative Three-Player Prisoners' Dilemma game. Players 1, 2, and 3 are involved in a crime and are interrogated separately. They can either choose to betray the others (Defect, D) or cooperate with them (Cooperate, C). The rewards and penalties are distributed based on the combination of choices made by the three players. 

The scoring is as follows:
- If all three choose to cooperate, they each get \{cooperate\} points.
- If all three choose to defect, they each get \{defect\} point.
- If one chooses to defect and the others choose to cooperate, the betrayer gets  \{one\_defect\} points and the collaborators each get 0 points.
- If two choose to defect and the other chooses to cooperate, the betrayers each get \{two\_defect\} points, and the collaborator gets 0 points.
The game is played for 5 rounds, and the total score is calculated at the end to determine the winner.
"""

&Please try to analyze how other players will choose to defect or corporate in the next round from your own perspective and others' perspectives. 
You must use the template:

In my opinion:
As \{player\_name\},
I think \{oth\_player1\} will xx, because...
I think \{oth\_player2\} will xx, because...
As for other players,
I think \{oth\_player1\} thinks:
\{player\_name\} will...
\{oth\_player2\} will...
I think \{oth\_player2\} thinks
\{player\_name\} will...
\{oth\_player1\} will...

&Based on your analysis of others and other's thoughts \- Remember to be strategic, establishing trust is to defect before others plan to do it to get the highest score. \- choose ``defect'' or ``cooperate'' to achieve the highest points for yourself.
You must reply with ``defect'' or ``cooperate''\\

\midrule
Public Good
&Welcome to the 'Public Good' game. Below are the game's settings and rules:
- Each player begins with 100 monetary points. Over \{game\_round\} rounds, you decide how many points to contribute to a communal pool at each round.
- After \{game\_round\} rounds of decision-making, the points in the communal pool will be multiplied by a factor of \{multiplier\} and distributed equally among all players.
- A player's final points are the sum of their remaining points and the shared points from the communal pool. The player who gets the highest final points wins the game.
- Every player must strategically invest their points to the communal pool to get more shared points and also be the one who invests the least to win the game.
- Usually, you can get more payback by investing more when the factor is larger.
&Please try to analyze whether other players will "reduce" or "increase" their contributions in the next round from your own perspective and others' perspective. 
- Remember, the payment in the communal pool is evenly shared by all players at the end of the game, so you need to make sure you invest the least money to get the highest repayment.
You must use the template:
In my opinion:
As \{player\_name\},
I think \{oth\_player1\} will xx, because...
I think \{oth\_player2\} will xx, because...
As for other players,
I think \{oth\_player1\} thinks:
\{player\_name\} will...
\{oth\_player2\} will...
I think \{oth\_player2\} thinks
\{player\_name\} will...
\{oth\_player1\} will...
&Based on your analysis of others and others' thoughts, make the decision about your own contribution to achieving the highest repayment for yourself. Remember
- Your total investment should be the least to win the game;
- Your target is to get the highest points and also promote the communal repayment to get as many points as possible at the end.
You must answer with the template ``I contribute xx''\\
    
\bottomrule
  \end{tabular}
  
  }
  \caption{Prompts in Cost-Sharing, Prisoner's  Dilemma, and Public Good.}
  \label{table:prompt-2}
\end{table*}

\end{document}